\documentclass[preprint,11pt]{article}
\usepackage{fullpage}

\usepackage{amsmath}
\usepackage{amssymb}
\usepackage{xspace}
\usepackage{dsfont}
\usepackage{algorithm}
\usepackage{algpseudocode}
\usepackage{mathtools}
\usepackage{dirtree}
\usepackage{multirow}
\usepackage{array}
\usepackage{cellspace}
\usepackage{hyperref}
\usepackage{url}
\usepackage{fancyvrb}
\usepackage{subfig}
\usepackage[dvipsnames]{xcolor}

\newcolumntype{M}[1]{>{\centering\arraybackslash}m{#1}}

\newenvironment{CodeChunk}{}{}
\DefineVerbatimEnvironment{CodeInput}{Verbatim}{fontshape=sl}

\newcommand{\ERT}{\operatorname{ERT}}

\newcommand{\R}{\ensuremath{\mathbb{R}}}

\newcommand{\N}{\ensuremath{\mathbb{N}}}

\newcommand{\oplga}{$(1,\lambda)$~GA\xspace}

\DeclareMathOperator{\ECDF}{ECDF}

\newcommand{\iohana}{\pkg{IOHanalyzer}\xspace}
\newcommand{\iohpro}{\pkg{IOHprofiler}\xspace}

\newcommand{\cascade}{\textcolor{gray}{$\blacktriangleright$}\xspace}

\let\proglang=\textsf
\newcommand{\pkg}[1]{{\selectfont #1}}

\begin{document}

\title{IOHanalyzer: Detailed Performance Analyses for Iterative Optimization Heuristics}

%
%
%
%
%
%

 \author{Hao Wang$^{2}$, Diederick Vermetten$^2$, Furong Ye$^2$, Carola Doerr$^1$, Thomas B\"ack$^2$}

 \date{$^1$Sorbonne Universit\'e, CNRS, LIP6, Paris, France\\
$^2$Leiden Institute of Advanced Computer Science, Leiden University, Leiden, The Netherlands}

\maketitle

\begin{abstract}
Benchmarking and performance analysis play an important role in understanding the behaviour of iterative optimization heuristics (IOHs) such as local search algorithms, genetic and evolutionary algorithms, Bayesian optimization algorithms, etc. This task, however, involves manual setup, execution, and analysis of the experiment on an individual basis, which is laborious and can be mitigated by a generic and well-designed platform. For this purpose, we propose IOHanalyzer, a new user-friendly tool for the analysis, comparison, and visualization of performance data of IOHs.

Implemented in \proglang{R} and \proglang{C++}, IOHanalyzer is fully open source. It is available on CRAN and GitHub. IOHanalyzer provides detailed statistics about fixed-target running times and about fixed-budget performance of the benchmarked algorithms with a real-valued codomain, single-objective optimization tasks. Performance aggregation over several benchmark problems is possible, for example in the form of empirical cumulative distribution functions. Key advantages of IOHanalyzer over other performance analysis packages are its highly interactive design, which allows users to specify the performance measures, ranges, and granularity that are most useful for their experiments, and the possibility to analyze not only performance traces, but also the evolution of dynamic state parameters. 

IOHanalyzer can directly process performance data from the main benchmarking platforms, including the COCO platform, Nevergrad, the SOS platform, and IOHexperimenter. An \proglang{R} programming interface is provided for users preferring to have a finer control over the implemented functionalities.
\end{abstract}

%



\sloppy{ 
\section{Introduction} 
\label{sec:intro}
Optimization problems not admitting exact solution approaches affect almost all aspects of our daily lives. They appear, for example, in product design, scheduling, data analysis, and {machine learning (e.g., hyper-parameter tuning). For instance, it is sometimes important to analyze the optimization procedure when training a neural network, which helps us understand the learning process.}
The intractability of these problems can have various reasons, e.g., a lack of problem-specific knowledge, limited access to problem data, or the inherent complexity of the underlying problem. Iterative optimization heuristics (IOHs) are algorithms designed to search for high-quality solutions of such problems. IOHs are characterized by a sequential structure, which aims to evolve good solutions by iteratively sampling the decision space. The distribution from which the solution candidates are sampled is adjusted after each iteration, to reflect the new information obtained from the last evaluations.  

IOHs are often randomized, both with respect to candidate generation and with respect to selecting the information stored from one iteration to the next. The optimization behavior of IOHs is therefore a highly complex system with many dependencies. This makes it very difficult to predict how well a particular IOH performs on a given problem. Existing theoretical results are limited to rather simple algorithms and/or problems, which are typically not representative for the complex strategies used in practice (see~\cite{DoerrN20,AugerD11,NeumannW10} for recent surveys of theoretical results). To gather a good understanding of the performance and the search behavior of realistic IOHs and applications, we are therefore often restricted to an empirical evaluation of these solvers, from which we may extrapolate accurate performance predictions. Supporting such empirical evaluations through a systematic experimental design is one of the primary goals of \emph{algorithm benchmarking}. Algorithm benchmarking addresses the selection of problem instances that are most suitable for an accurate performance extrapolation, the experimental setup of the data generation, the choice of the performance indicators and their visualizations, the choice of the statistics used to compare two or more algorithms, etc. In practice, those various aspects of algorithm benchmarking make it laborious and demanding for researchers to handle the details of experimentation, which calls for a standard and easy-to-use software implementation of algorithm benchmarking that would drastically reduce the manual work for practitioners.

\subsection{IOHanalyzer: Overview and Availability}
\label{sec:intro:tool}

In this work, we present \iohana, a versatile, user-friendly, and highly interactive platform for the assessment, comparison, and visualization of IOH performance data. \iohana is designed to assess the empirical performance of sampling-based optimization heuristics in an algorithm-agnostic manner. Our \textbf{key design principles} are 1) an easy-to-use software interface, 2) interactive performance analysis, and 3) convenient export of reports and illustrations.

\begin{figure}[t]
\centering
\includegraphics[width=\textwidth]{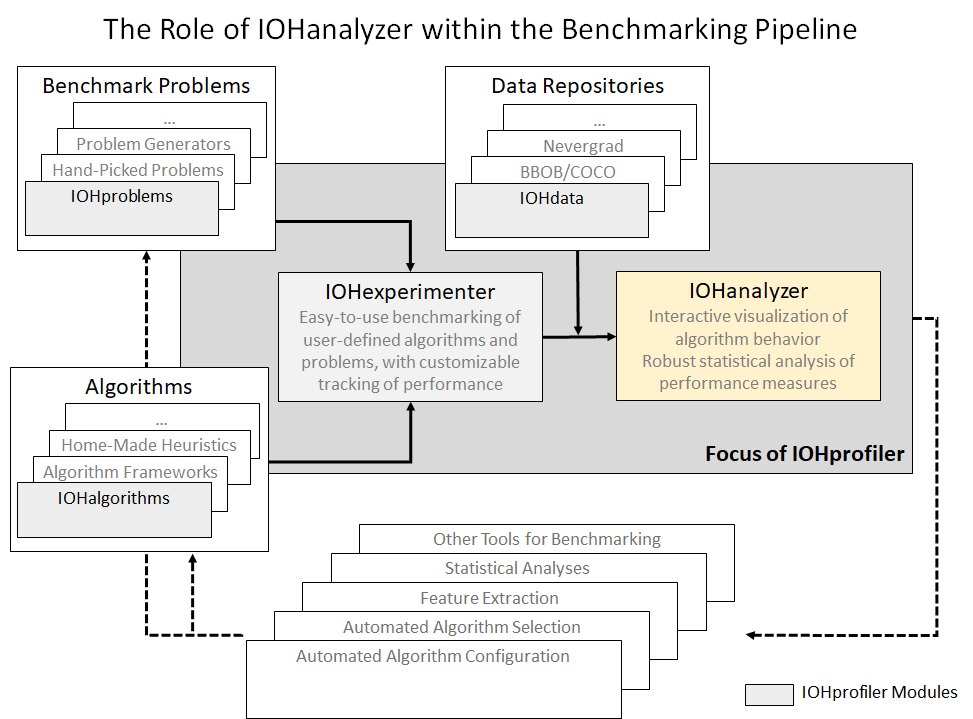}
\caption{
\iohana is a module of the \iohpro benchmarking environment, which targets different steps of the benchmarking pipeline.}
\label{fig:pipeline}
\end{figure}

\iohana is developed as the data analysis component of \iohpro, a benchmarking platform that aims to integrate various elements of the entire benchmarking pipeline, ranging from problem (instance) generators and modular algorithm frameworks over automated algorithm configuration techniques and feature extraction methods to the actual experimentation, data analysis, and visualization~\cite{IOHprofiler}. An illustration of the interplay between these different components is provided in Figure~\ref{fig:pipeline}. 
Notably, \iohpro already provides the following components: 
\begin{itemize}
    \item \textbf{IOHproblems:} a collection of benchmark problems. This component currently comprises (1) the PBO suite of pseudo-Boolean optimization problems suggested in~\cite{DoerrYHWSB20}, (2) the 24 numerical, noise-free BBOB functions from the \underline{CO}mparing \underline{C}ontinuous \underline{O}ptimizers (COCO) platform~\cite{hansen2016coco}, and (3) the Wmodel problem generator proposed in~\cite{Weise:2018}. 
    \item \textbf{IOHalgorithms:} a collection of IOHs. For the moment, the algorithms used for the benchmark studies presented in~\cite{DoerrYHWSB20,AmineGECCO,modCMAGECCO} are available. 
    This subsumes textbook algorithms for pseudo-Boolean optimization, an integration to the object-oriented algorithm design framework ParadisEO~\cite{ParadisEO}, and the modular algorithm framework for CMA-ES variants originally suggested in~\cite{modular-CMAES} and extended in~\cite{modCMAGECCO}. Further extensions for both combinatorial and numerical solvers are in progress. 
    \item  \textbf{IOHdata:} a data repository for benchmark data. This repository currently comprises the data from the experiments performed in~\cite{DoerrYHWSB20}, a sample data set used in this paper, and some selected data sets from the COCO repository~\cite{BBOBdata}. \textbf{IOHdata} also contains performance data from Facebook's Nevergrad benchmarking environment~\cite{nevergrad}, which can be fetched from their repository upon request. 
    \item \textbf{IOHexperimenter:} the experimentation environment that executes IOHs on \textbf{IOHproblems} or external problems and automatically takes care of logging the experimental data. It allows for tracking the internal parameter of IOHs and supports various customizable logging options to specify when to register a data record.
    \item \textbf{IOHanalyzer:} the data analysis and visualization tool presented in this work. 
\end{itemize}

\iohana takes as input benchmarking data sets, generated, e.g., by IOHexperimenter, through the COCO platform, or 
through the Nevergrad environment. 
Of course, users can also use their own experimentation platform (the formatting requirements for the input files are described in Appendix~\ref{subsec:data-format}). \iohana provides an evaluation platform for these performance traces, which allows users to choose the performance measures, the ranges, and the precision of the displayed data according to their needs. In particular, \iohana supports both a fixed-target and a fixed-budget perspective, and allows various ways of aggregating performances across different problems (or problem instances). In addition to these performance-oriented analyses, \iohana also offers statistics about the evolution of non-static algorithmic components, such as, for example, the hyperparameters suggested by a self-adjusting parameter control scheme. These features will be described in more detail in Section~\ref{sec:gui}, where the reader can also find illustrated examples. 

The \proglang{R} programming interface of \iohana offers a fine control on the data and functionalities implemented therein. \iohana is written in \proglang{R} and \proglang{C++} and makes use of the two R packages \pkg{plotly}~\cite{plotly} and \pkg{shiny}~\cite{shiny}. {The version of the software described in this paper is \emph{v0.1.6.1}, which has been made available on zenodo~\cite{zenodo_iohanalyzer}. This repository also contains all datasets used for the examples illustrated in this paper}.  
For users less experienced with programming in R we offer a web-based graphical user interface (GUI), to which users can load their own data or use data from the \textbf{IOHdata} repository. 
 
The stable release of the \iohana package is distributed through CRAN (\url{https://CRAN.R-project.org/package=IOHanalyzer}). It can be easily installed in an \proglang{R} console:
\begin{CodeChunk}
\begin{CodeInput}
R> install.packages("IOHanalyzer")
\end{CodeInput}
\end{CodeChunk}
The latest version is hosted on GitHub (\url{https://github.com/IOHprofiler/IOHanalyzer}, part of the IOHprofiler project), which can be installed using the \pkg{devtools} library as follows:\footnote{The GitHub-page gets updated more frequently with minor changes, while the CRAN-version is generally only updated only when major modifications are made.} 
\begin{CodeChunk}
\begin{CodeInput}
R> devtools::install_github("IOHprofiler/IOHanalyzer")
\end{CodeInput}
\end{CodeChunk}
An up-to-date  documentation is maintained on the wiki page, available at~\url{https://iohprofiler.github.io/}. The web-based GUI of \iohana is hosted at~\url{http://iohprofiler.liacs.nl}.

The first use case of \iohana was the comparison of different variants of the $(1+\lambda)$~evolutionary algorithm (EA)~\cite{DoerrYRWB18}. A number of improvements were made subsequently, and the first study of an important number experiments was reported in~\cite{DoerrYHWSB20}. In the meantime, \iohana has been used in a number of studies, including~\cite{HoreshBS19,YeDB19,CalvoSCD0BL19}. It is under constant development. Some of the major ongoing extensions will be discussed in Section~\ref{sec:outlook}.

\subsection{Related Benchmarking Environments}

As argued above, benchmarking IOHs is an essential task towards a better understanding of IOHs. It is therefore not surprising that a large number of different tools have been developed for this purpose. For reasons of space, we can only summarize a few of them and concentrate on those which come closest to \iohana in terms of functionality and scope. 

In evolutionary computation, the arguably best established benchmarking environment is the already mentioned COCO  platform~\cite{hansen2016coco}. Originally designed to compare derivative-free optimization algorithms operating on numeric optimization problems~\cite{Comparing31}, the tool has seen several extensions in the last years, e.g., towards multi-objective optimization~\cite{COCOemo}, mixed-integer optimization~\cite{COCOmip}, and large-scale optimization~\cite{bbob-large-ASOC}. COCO consists of an experimentation part that produces data files with detailed performance traces, and an automated data analysis part in which a fixed number of standardized analyses are automatically generated. The by far most reported performance measures from the COCO framework are \emph{empirical cumulative distribution function} (ECDF) curves, see Section~\ref{sec:background} for definitions. The COCO software has a strong focus on \emph{fixed-target performances}~\cite{COCOperformance}, i.e., on the time needed to find a solution of a certain quality. 

COCO has been a major source of inspiration for the development of IOHprofiler. What concerns the performance assessment, the key difference between COCO and our \iohana is in the {interactive interface that allows users of \iohana to study different performance measures, to change their ranges, and granularity}. As mentioned, COCO performance files can be conveniently analyzed by \iohana.

Another important software environment for benchmarking sampling-based optimization heuristics is the \pkg{Nevergrad} framework~\cite{nevergrad}. As with COCO, Nevergrad implements functionalities for both experimentation and performance analysis, accommodating continuous, discrete, and mixed-integer problems. It has a strong focus on noisy optimization, but also comprises several noise-free optimization problems. In addition to studying IOHs, Nevergrad has a special suite to compare one-shot optimization techniques, i.e., non-iterative solvers. The current focus of Nevergrad is to be seen on the problem side, as it offers several new benchmark problems, such as the structured optimization problems which are aggregated in their own test suite. Nevergrad also provides interfaces to the following benchmark collections:  LSGO~\cite{lsgo}, YABBOB~\cite{versatile}, Pyomo~\cite{pyomo}, MLDA~\cite{mlda}, and MuJoCo~\cite{mujoco}. 
The performance evaluation, however, is much more basic than those of COCO or \iohana, in that only the quality of the finally recommended point(s) is stored, but no information about the search trajectory. That is, apart from taking a \emph{fixed-budget perspective}, Nevergrad does not store performance traces, but only the final output. \iohana can interpret and visualize the csv files produced by Nevergrad. An extension of Nevergrad to allow for the same tracking features as \iohana is currently under construction, in a joint collaborative effort between the Nevergrad and the \iohana development teams. 

Focusing on the algorithm design task, HeuristicLab~\cite{HeuristicLab} provides a relatively large collection of various IOHs (e.g., population-based search algorithms) as well as machine learning algorithms (e.g., Support Vector Regression), which are represented as graphs of operators. In HeuristicLab, new algorithms can be constructed by combining existing operators in a graphical user interface, avoiding the laborious coding details. While IOHprofiler mainly targets the black-box optimization problem, HeuristicLab incorporates a very diversified set of benchmark problems, ranging from the symbolic regression to data analysis problems. It implements the parallel execution of algorithms for the ease of benchmarking. In contrast to IOHanalyzer, which is available via a web interface and contains many detailed performance analyses and interactive plots, HeuristicLab is distributed via platform-dependent applications and includes some basic static plots (e.g., box plots) for assessing the empirical performance from a \emph{fixed-budget perspective}. 

Several other tools have been developed for displaying performance data and/or the search behavior in decision space. However, all tools that we are aware of allow much less flexibility with respect to the performance measures, the ranges, and the granularity of the analysis or focus on selected aspects of performance analysis only (e.g., \cite{Calvo:2018:BIA:3205651.3205658,EFTIMOV2019255} study statistical significance, whereas~\cite{eaf,SchapermeierGK21} aim to visualize performance with respect to multiple objectives). The ability of \iohana to link the evolution of algorithms' parameters to the evolution of solutions' quality seems to be unique. 

\section{Background} 
\label{sec:background}

This section provides the background and motivation for developing 
\iohana. In particular, we discuss black-box problems and their optimization 
and we recall the most relevant performance indicators that will be used in subsequent sections. 

\subsection{Iterative Optimization Heuristics}
\label{subsec:optimization}

\begin{algorithm}[t]
	\caption{Blueprint of an iterative optimization heuristic (IOH) optimizing a function $f:\mathcal{S} \rightarrow \R$.}
	\label{alg:IOH}
	\begin{algorithmic}[1]
	\Procedure{\textsc{ioh}}{}
	\State{$t \leftarrow 0$} \Comment{iteration counter}
	\State{$\mathcal{H}(0) \leftarrow \emptyset$} \Comment{search history information}
	\State{choose a distribution $\Lambda(0)$ on $\N$} \Comment{distribution of the number of samples}
	\While{termination criterion not met}
	        \State $t \leftarrow t+1$ 
            \State sample $\lambda(t) \sim \Lambda(t-1)$  \Comment{nbr. of points to be evaluated}
	        \State Based on $\mathcal{H}(t-1)$ choose a distribution $D(t)$ on $\mathcal{S}^{\lambda(t)}$	\Comment{choice of sampling distribution}
			\State{sample $\left(x^{(t,1)}, \ldots, x^{(t,\lambda(t))}\right) \sim D(t)$} \Comment{candidate generation}
			\State evaluate $f\left(x^{(t,1)}\right), \ldots, f\left(x^{(t,\lambda(t))}\right)$ \Comment{function evaluation}
	    	\State choose $\mathcal{H}(t)$ and $\Lambda(t)$  \Comment{information update}
	\EndWhile
	\EndProcedure
	\end{algorithmic}
\end{algorithm} 

We study the optimization of problems of the type $f\colon \mathcal{S} \rightarrow \mathbb{R}$, i.e., we assume our problem to be a single-objective, real-valued objective function {(i.e., problems for which the quality of possible solutions is rated by real numbers)}, defined over a search space $\mathcal{S}$. We do not make any assumption on the set $\mathcal{S}$; it can be discrete or continuous, constrained or unconstrained. 
We do not require that $f$ is explicitly modeled, i.e., $f$ can very well be a \emph{black-box optimization problem}
{, i.e., a problem for which we are able to evaluate the quality of points $x \in \mathcal{S}$ (e.g., through computer simulations or through physical experiments) but for which we do not know the mapping $x \mapsto f(x)$ without performing such evaluations.}  
Intermediate \emph{gray-box} settings are also possible, {i.e., problems for which \emph{some} information about the mapping $x \mapsto f(x)$ is available (see, for example, the discussion in~\cite{WhitleyCG16}, where a setting is analyzed in which users have information about the interaction between different variables)}.  
To ease notation, we nevertheless speak of black-box optimization in such cases, i.e., even when some a priori information about the problem $f$ is available. We emphasize that the sampling-based optimization algorithms studied in our work can be competitive even when the problem $f$ is explicitly known. The low auto-correlation binary sequence (LABS) problem is a good example for such a problem that can be defined in two lines, but for which the best known solvers are sampling-based~\cite{LABS_Packebusch2016}. {The only important feature of the performance traces that can be analyzed by \iohana is that they rely on the evaluation of possible solution candidates -- regardless of how these have been created.} 

For convenience of presentation, we consider in this document \textbf{maximization} as objective. Note, though, that \iohana automatically detects whether minimization of maximization is considered, and adjusts the plots and statistics accordingly. For example, the COCO and Nevergrad data sets typically consider minimization, whereas the PBO suite of \textbf{IOHproblems} studies maximization.

The class of algorithms that we are interested in are \emph{Iterative Optimization Heuristics} (IOHs). IOHs are entirely sampling-based, i.e., they sample the search space $\mathcal{S}$ and use the function values $f(x)$ of the evaluated samples $x$ to guide the search. Algorithm~\ref{alg:IOH} provides a blueprint for IOHs. Classical examples for IOHs are deterministic and stochastic local search algorithms (this class includes Simulated Annealing~\cite{SA83} and Threshold Accepting~\cite{TA90Dueck} as two prominent examples), genetic and evolutionary algorithms~\cite{EibenS15}, Bayesian Optimization and related global optimization algorithms~\cite{Jones2001}, Estimation of Distribution algorithms~\cite{EDA-book}, and Ant Colony Optimization algorithms~\cite{ants-book}.

\subsection{Selected Performance Indicators}
\label{subsec:performance}
Unlike in optimization scenarios in which problem data is accessible without function querying and where solutions are hence typically generated constructively (as opposed to the sampling-based approach taken by IOHs), 
the most commonly studied performance measures in black-box optimization are based on the number of function evaluations. That is, instead of counting arithmetic operations or CPU time, we measure performance by counting the number of function evaluations that are performed to reach a certain quality threshold (\emph{fixed-target setting}) or we measure the quality of the best found solution that could be recommended after a certain budget of function evaluations has expired (\emph{fixed-budget setting}). {Measuring the performance in the number of function evaluations is a classic assumption made in the black-box optimization literature~\cite{COCOperformance}. In contrast to CPU time, this measure is machine-independent and not (or at least much less) sensitive with respect to the actual implementation.}

As discussed above, many state-of-the-art IOHs are randomized in nature, therefore yielding random performance traces even when the underlying problem $f$ is deterministic. The performance space is spanned by the number of evaluations, by the quality of the assessed solutions, and by the probability that the algorithm has found within a given budget of function evaluations a solution that is at least as good as a given quality threshold.

\paragraph{Basic Notation} 
To define the performance measures covered by \iohana we use the following notation. 
\begin{itemize}
	\item $\mathcal{F}$ denotes the set of problems under consideration. Each problem (or problem instance, depending on the context) $f \in \mathcal{F}$ is assumed to be a function $f\colon \mathcal{S} \rightarrow \mathbb{R}$. The \emph{dimension} of $\mathcal{S}$ is denoted by $d$. We often consider scalable functions that are defined for several or all dimensions $d \in \N$. In such cases, we make the dimension explicit. 
	\item $\mathcal{A} = \{A_1, A_2, \ldots\}$ is the set of algorithms under consideration. $\mathcal{A}$ can be finite or infinite. Often, $\mathcal{A}$ is a configurable meta-algorithmic framework, which allows users to specify parameters such as the degree of parallelism, the intensity of the local perturbations, the memory size, the use (or not) of recombination operators, etc.
	\item We denote by $r$ the number of independent runs of an algorithm $A\in\mathcal{A}$ on problem $f\in\mathcal{F}$ in dimension $d$.
	\item $T(A,f,d,B,v,i) \in \N \cup \{\infty\}$ is a \emph{fixed-target measure.} It denotes the number of function evaluations that algorithm $A$ performed, in its $i$-th run and when maximizing the $d$-dimensional variant of problem $f$, to find a solution $x$ satisfying $f(x) \ge v$. When $A$ did not succeed in finding such a solution within the maximal allocated budget $B$, $T(A,f,d,B,v,i)$ is set to $\infty$. Several ways to deal with such failures are considered in the literature, as we shall discuss in the next paragraphs. 
	\item Similarly to the above, $V(A,f,d,t,i) \in \R$ is a \emph{fixed-budget measure.} It denotes the function value of the best solution that algorithm $A$ evaluated within the first $t$ evaluations of its $i$-th run, when maximizing the $d$-dimensional variant of problem $f$. 
\end{itemize}

\paragraph{Descriptive Statistics} We next recall some basic descriptive statistics.
\begin{itemize}
	\item The average function value given {a budget value $t$} is simply 
	$$ \bar{V}(t)=\bar{V}(A,f,d,t)=\frac{1}{r}\sum_{i=1}^r V(A,f,d,t,i). $$
	As we do with all other measures, we omit explicit mention of  $A$, $f$, and $d$ when they are clear from the context. 
	\item The \emph{Penalized Average Runtime} (PAR-$c$ score{, where {$c\geq1$} is the penalty factor}) for a given target value $v$ is defined as
	\begin{equation}
	\operatorname{PAR-c}(v)=\operatorname{PAR-c}(A,f,d,B,v)=\frac{1}{r}\sum_{i=1}^r \min\left\{T(A,f,d,B,v,i), cB\right\}, \label{eq:sample-mean}
	\end{equation}
	i.e., the PAR-$c$ score is identical to the sample mean when all runs successfully identified a solution of quality at least $v$ within the given budget $B$, whereas non-successful runs are counted as $cB$. In \iohana, we typically study the PAR-1 score, which, in abuse of notation, we also refer to as the mean.
	\item Apart from mean values, we are often interested in quantiles, and in particular in the \emph{sample median} of the $r$ values $\left\{T(A,f,d,B,v,i\}\right)_{i=1}^r$ and $\left\{V(A,f,d,{t},i\}\right)_{i=1}^r$, respectively. By default, \iohana calculates the $2\%, 5\%, 10\%, 25\%, 50\%, 75\%, 90\%, 95\%$, and $98\%$ percentiles (denoted as $Q_{2\%}, Q_{5\%},\ldots, Q_{98\%}$) for both running times and function values. 
	\item We also study the \emph{sample standard deviation} of the running times and function values, respectively.
	\item The \emph{empirical success rate} is the fraction of runs in which algorithm $A$ reached the given target $v$ {within the maximal number $B$ of allowed function evaluations}. That is, in the case of a maximization problem, 
	\begin{equation}\label{eq:success-rate}
		\widehat{p}_s = \widehat{p}_s (A,f,d,B,v) =
		{\frac{1}{r}\sum_{i=1}^r\mathds{1}(V(A,f,d,B,i) >v)}
 		= \frac{1}{r}\sum_{i=1}^r\mathds{1}(T(A,f,d,B,v,i) < \infty), 
	\end{equation}
	where $\mathds{1}(\mathcal{E})$ is the characteristic function of the event $\mathcal{E}$. 
	\end{itemize}

\paragraph{Expected Running Time} 
An alternative to the PAR-$c$ score is the expected running time (ERT). ERT assumes independent restarts of the algorithm whenever it did not succeed in finding a solution of quality at least $v$ within the allocated budget $B$. Practically, this corresponds to sampling indices $i \in \{1,\ldots, r\}$ (i.i.d.~uniform sampling with replacement) until hitting an index $i$ with a corresponding value $T(A,f,d,B,v,i) < \infty$. The running time would then have been $m B + T(A,f,d,B,v,i)$, where $m$ is the number of sampled indices of unsuccessful runs. The average running time of such a hypothetically restarted algorithm is then estimated as  
\begin{align}
\ERT(A,f,d,B,v) 
	& = \frac{\sum_{i=1}^{r} \min\left\{T(A,f,d,B,v,i), B\right\}}{r \widehat{p}_s} \nonumber \\
	& = \frac{\sum_{i=1}^{r} \min\left\{T(A,f,d,B,v,i), B\right\}}{\sum_{i=1}^r\mathds{1}(T(A,f,d,B,v,i) < \infty)}. \label{eq:ERT}
\end{align}
Note that ERT can take an infinite value when none of the runs was successful in identifying a solution of quality at least $v$. 

\paragraph{Cumulative Distribution Functions} For the fixed-target and fixed-budget analysis, \iohana estimates probability density (mass) functions and computes empirical cumulative distribution functions (ECDFs). For the fixed-budget function value, its probability density function is estimated via the well-known Kernel Density Estimation (KDE) method, which approximates the density function by a superposition of kernel functions (e.g., Gaussian functions with a fixed width) centered at each data point~\cite{HastieTF09}. Intuitively, a set of crowded data points would lead to a very peaky empirical density due to massive superpositions of the kernel, while a set of distant points can only generate a relatively flat curve. For the fixed-target running time (an integer-valued random variable), we estimate its probability mass function by treating it as a real value and applying the KDE method.
For a set $\{T(A,f,d,v,i)\}_{i=1}^r$ of fixed-target running times, its ECDF is defined as the fraction of runs which successfully found a solution of quality at least $v$ within a budget of at most $t$ function evaluations. That is, 
$$\ECDF(A,f,d,v,t)=\frac{1}{r}\sum_{i=1}^r \mathds{1}(T(A,f,d,v,i) \leq t).$$
ECDF values are most typically used in aggregated form. \iohana uses the following two aggregations:
%
\begin{itemize}
	\item The aggregation over a set $\mathcal{V}$ of \emph{target values}:
	\begin{equation}
		\ECDF(A,f,d,\mathcal{V},t) = \frac{1}{r|\mathcal{V}|}\sum_{v\in \mathcal{V}}\sum_{i=1}^{r} \mathds{1}(T(A,f,d,v,i) \leq t),  \label{eq:ECDF-agg-target}
	\end{equation}  
	i.e., the fraction of (run,target value) pairs $(i,v)$ for which algorithm $A$ has identified a solution of quality at least $v$ within a budget of at most $t$ function evaluations.
	\item Given a set of functions $\mathcal{F}$ and {a mapping $\mathcal{V}\colon \mathcal{F} \rightarrow 2^\mathbb{R}$ that specifies the target values to consider for each function}, the ECDF can be further aggregated by {the following definition:} 
	\begin{equation}
	{
	\ECDF(A,\mathcal{F},d,\mathcal{V},t) = \frac{1}{r\sum_{f\in \mathcal{F}}|\mathcal{V}(f)|}\sum_{f\in\mathcal{F}}\sum_{v\in \mathcal{V}(f)}\sum_{i=1}^{r} \mathds{1}(T(A,f,d,v,i) \leq t). } \label{eq:ECDF-agg-fct}
	\end{equation} 	
\end{itemize}
The aggregated ECDFs for function values $V(A,f,d,t,i)$ can be defined in the similar manner. By default, \iohana can generate the targets in a linear or log-linear way, as well as the predefined targets commonly used in the COCO framework. However, all of these targets can be changed by the user. 

\section{Graphical User Interface}
\label{sec:gui}

The web-based Graphical User Interface (GUI) may be the most convenient access to \iohana for users who are not sufficiently familiar with programming in \proglang{R}, as well as for users who are more interested in comparing (with) data from the existing data sets collected in the performance data repository \textbf{IOHdata}. 
In this and the next section we use an exemplary data set called ``sample\_data'' prepared for this article, which comprises selected performance data from the study presented in~\cite{DoerrYHWSB20}. 
This data set is already available in the web-based GUI and the user can load it from the ``Load Data from Repository'' box therein (see the bottom right part in Figure~\ref{fig:GUI-data-loading}). 
More precisely, we have selected from this data set the performance files for two algorithms ({Randomized Local Search} (RLS) and {the Genetic Algorithm (GA) variant} \oplga, see~\cite{DoerrYHWSB20} for a detailed description and references) on four problems in two dimensions $d\in\{16, 100\}$. All problems analyzed in~\cite{DoerrYHWSB20} are of the type $f:\{0,1\}^n \rightarrow \R$, and both the problem suite as well as the dataset are named PBO (for \emph{pseudo-Boolean optimization}) in \textbf{IOHproblems} and \textbf{IOHdata}, respectively. 
To use this data set locally, users need to create a folder named \verb|repository/sample_data| under the home directory (i.e., $\sim$\verb|/repository/sample_data|), download the data set from \url{https://github.com/IOHprofiler/IOHdata/blob/master/sample\_data.rds}, and move the data set to this location.

The \iohana GUI is invoked through the following commands: 

\begin{CodeChunk}
\begin{CodeInput}
R> library(IOHanalyzer)
R> runServer()
Loading required package: shiny

Listening on 127.0.0.1:3943
\end{CodeInput}
\end{CodeChunk}

\begin{figure}[!ht]
\centering
\includegraphics[width=\textwidth, trim=0mm 00mm 0mm 0mm, clip]{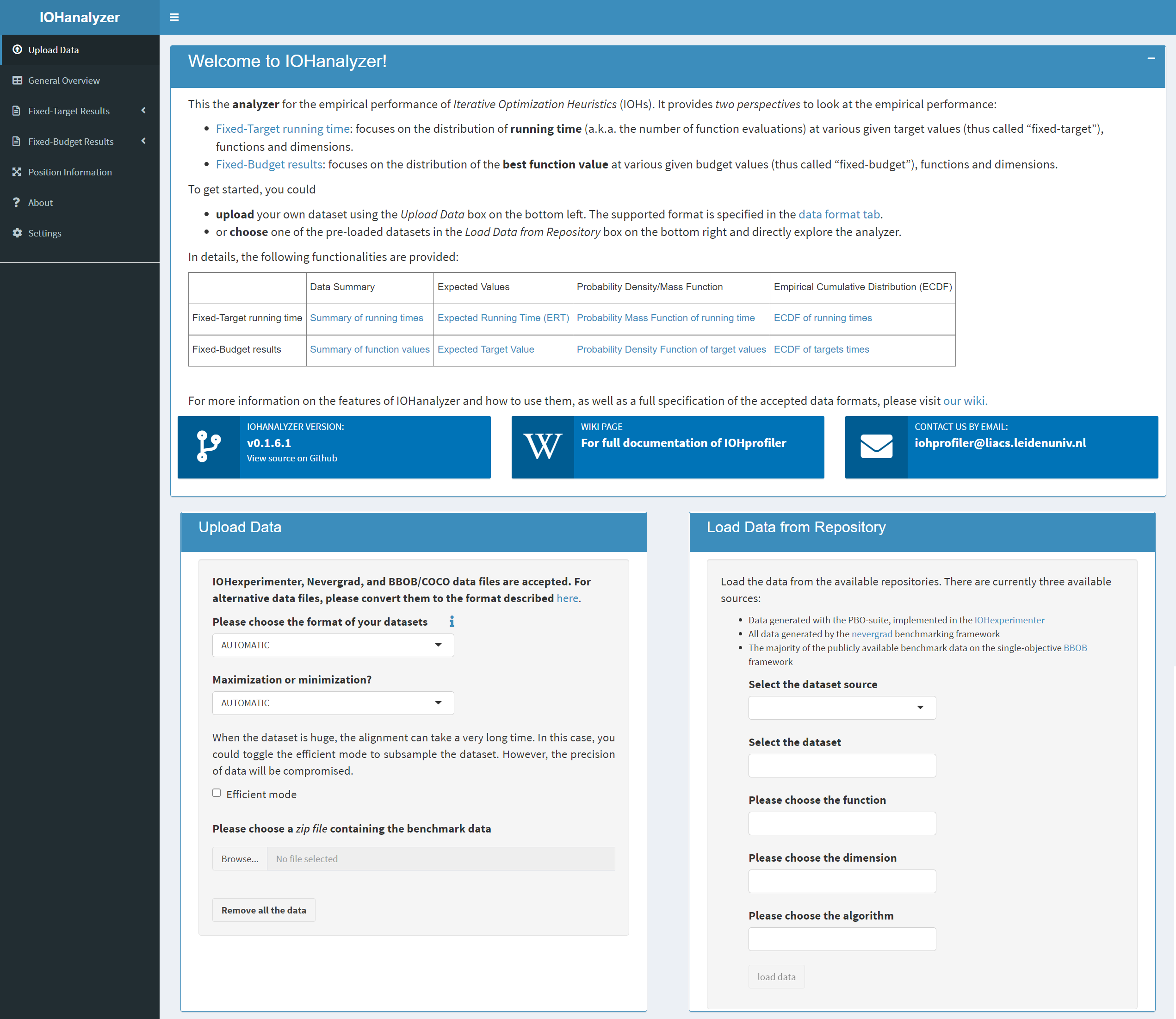}
\caption{\label{fig:GUI-data-loading} Screenshot of the GUI immediately after launching the GUI server. Some general information, such as the current version and relevant links are provided at the top. The user can choose a data set from our online data repository in the drop-down menu of right column, or upload local data using the column on the left.}
\end{figure}

This will start the GUI server on the local machine (hence using IP address 127.0.0.1) and a random port number. The web browser will be launched and connect to this address immediately after starting the server. The screenshot of the ``welcome page'' is shown in Figure~\ref{fig:GUI-data-loading}. The performance statistics are arranged in four major sections, which can be chosen in the side menu on the left. The side menu is organized as follows. 
\begin{enumerate}
	\item \textbf{Upload Data:} In this section users can upload their own performance data files and/or choose the data from the repository against which the data shall be compared. The format of the data files which can be uploaded is discussed in Appendix~\ref{subsec:data-format}. 
	\item \textbf{General Overview:} On this tab, we show a summary of algorithms, function/problems, dimensions, the number of runs, and the best reached function values per function and algorithm appearing in the data set loaded by the user.
	\item \textbf{Fixed-Target Result:} This section covers the fixed-target performance statistics summarized in Table~\ref{tab:summary-iohanalyzer-fixed-target}. A detailed description will be given in Section~\ref{subsec:fixed-target-results}. 
	\item \textbf{Fixed-Budget Results:} This section covers the fixed-budget statistics summarized in Table~\ref{tab:summary-iohanalyzer-fixed-budget}. 
    \item \textbf{Position Information:} A parallel coordinate plot allows the user to display the final point resulting from each run of an algorithm. This can be used for comparing the distribution of the final solutions found across many algorithms.
	\item \textbf{About:} A concise description of the \iohana and installation guide are included here, together with information on the development team, the license, and acknowledgements.  
	\item \textbf{Settings:} Here, the user can change the color schemes, the font size, and the image size used in plotting and other general settings controlling the calculation of descriptive statistics. 
\end{enumerate}

In general, the interactive plotting (enabled by the \pkg{plotly} library) is turned on by default, displaying more details in the plot when the user hovers the mouse over it, e.g., the value of a curve at the mouse cursor. The interactive plotting also allows the user to zoom in/out and to hide/show a curve from some algorithms, which will be helpful when many algorithms are rendered simultaneously. Also, all plots can be downloaded in the following formats: \texttt{pdf, png, eps}, and \texttt{svg}. Most data tables can be downloaded in csv format. 

\subsection{The ``Upload Data'' Section}
\label{subsec:upload}
The GUI interface to load the experimental data is shown in Figure~\ref{fig:GUI-data-loading}. in which the user is asked to upload a \emph{compressed archive}. The following compression formats are supported: \texttt{*.zip}, \texttt{*.bz}, \texttt{*.tar}, \texttt{*.xz}, \texttt{*.gz} and \texttt{*.rds} (previously processed . Note that, when the user's data set is very large to handle, 
it is possible to speed up the uploading (and hence plotting) procedure by toggling option \verb|Efficient mode| on, in which the original data set is downsampled uniformly at random. Note that the data-uploading module will automatically detect whether maximization or minimization has been the objective, given the uploaded data set follows the formatting requirements described in Appendix~\ref{subsec:data-format}.

When using the online version of GUI (\url{http://iohprofiler.liacs.nl/}), the user can also load the data sets from \textbf{IOHdata}, using the ``Load Data from Repository'' box on the right (see Figure~\ref{fig:GUI-data-loading}). 

After loading the data, \iohana will prompt a summary table of loaded data sets in the ``List of Processed Data'' box on the bottom of the page (not shown in Figure~\ref{fig:GUI-data-loading}). This allows users to check if the data loading process has been performed correctly.

\subsection{The ``Fixed-Target Results'' Section}
\label{subsec:fixed-target-results}

In the fixed-target section, the user can analyze the number of function evaluations that the algorithms performed before finding for the first time a solution meeting a certain quality criterion. This section has two main subsections, one for the performance evaluation of a \emph{single function} and one for the evaluation of performance data for \emph{multiple functions}. Table~\ref{tab:summary-iohanalyzer-fixed-target} summarizes the main fixed-target performance statistics that \iohana offers.  

\begin{table}[!htbp]
\scriptsize
\centering
\renewcommand{\arraystretch}{1.3}
\begin{tabular}{|M{.6cm}|m{2cm}|m{3cm}|m{6cm}|}
\hline
Section & Group &  Functionality  & Description \\ \hline\hline
\multirow{20}{*}{\rotatebox[origin=r]{90}{Single Functions}} & \multirow[c]{5}{1.5cm}{Data Summary}  &\emph{Data Overview} & The best, worst, mean, median values and success rate of selected algorithms.  \\ \cline{3-4}
& & \emph{Runtime Statistics} & The mean, median, quantiles, success rate and ERT at an evenly spaced sequence of targets controlled by $f_{\text{min}},f_{\text{max}}$ and $\Delta f$. \\ \cline{3-4}
& & \emph{Runtime Samples} & The running time sample at an evenly spaced sequence of targets controlled by $f_{\text{min}},f_{\text{max}}$ and $\Delta f$.  \\\cline{2-4}
& \multirow[c]{3}{1.5cm}{Expected Runtime} & \emph{ERT: single function} & The progression of ERT over targets, whose range is controlled by the user. \\ \cline{3-4}
& & {\emph{Expected Runtime Comparisons}} & {Comparing the ERT values of selected algorithms at pre-computed targets across all problem dimensions on a chosen problem.}  \\ \cline{2-4}
& \multirow[c]{3}{1.5cm}{Probability Mass Function} & \emph{Histogram} & The histogram of the running time at a target specified by 
the user on \textbf{one} function. \\\cline{3-4}
& & \emph{Probability Mass Function} & The probability mass function of the running time at a target specified by  
the user on \textbf{one} function. \\\cline{2-4}
& \multirow[c]{2}{1.0cm}{Cumulative Distribution}  & \emph{ECDF: single target}& On \textbf{one} function, the ECDF of the running time at \textbf{one} target specified by 
the user. \\ \cline{3-4}
& & \emph{ECDF: single function}& On \textbf{one} function, ECDFs aggregated over \textbf{multiple} targets. \\ \cline{3-4}
\cline{2-4}
& \multirow[c]{5}{1.5cm}{Algorithm Parameters}  & \emph{Expected Parameter Value} & The progression of expected value of parameters over \textbf{targets}, whose range is controlled by the user. \\ \cline{3-4}
& & \emph{Parameter Statistics} & The mean, median, quantiles of recorded parameters at an evenly spaced sequence of targets controlled by $f_{\text{min}},f_{\text{max}}$ and $\Delta f$. \\ \cline{3-4}
& & \emph{Parameter Sample} & The sample of recorded parameters at an evenly spaced sequence of targets controlled by $f_{\text{min}},f_{\text{max}}$ and $\Delta f$. \\ \cline{2-4} 
& Statistics  & \emph{Hypothesis Testing} & The two-sample Kolmogorov-Smirnov test applied on the running time at a target value for each pair of algorithms. A partial order among algorithms is obtained from the test. \\ \cline{1-4}
\multirow{19}{*}{\rotatebox[origin=r]{90}{Multiple Functions}} & \multirow[c]{2}{1.5cm}{{Data Summary}} & \emph{Multi-Function Statistics
} & Descriptive statistics for all functions at a single target value. \\ \cline{3-4}
& & \emph{Multi-Function Hitting Times
} & Raw hitting times for all functions at a single target value.\\ \cline{2-4}
& \multirow[c]{3}{1.5cm}{Expected Runtime} & \emph{ERT: all functions} & The progress of ERTs are grouped by functions and the range of targets are automatically determined. \\ \cline{3-4}
& & \emph{Expected Runtime Comparisons} & The ERTs at the best target found on each function (one fixed dimension) is plotted against the function ID for each algorithm.\\ \cline{2-4}
& \parbox[c][0.8cm]{1.2cm}{Cumulative \\Distribution} & \parbox[c][0.8cm]{3cm}{\emph{ECDF: all functions}} & \parbox[c][0.8cm]{6cm}{On \textbf{all} functions, ECDFs aggregated over \textbf{multiple} targets.} \\ \cline{2-4}
& \multirow[c]{4}{2.5cm}{{Deep Statistics}} & \emph{Ranking per Function} & Per-function statistical ranking procedure from the Deep Statistical Comparison Tool (DSCTool)~\cite{EftimovPK20}.\\ \cline{3-4}
& & \emph{Omnibus Test} & Use the results of the per-function ranking to perform an omnibus test using DSC. \\ \cline{3-4}
& & \emph{Posthoc comparison} & Use the results of the omnibus test to perform the post-hoc comparison. \\ \cline{2-4}
& Ranking & \emph{Glicko2-based ranking} & For each pair of algorithms, a running time value at a given target is randomly chosen from all sample points in each round of the comparison. The glicko2-rating 
is used to determine the overall ranking from all comparisons. \\ \cline{2-4}
& {Portfolio} & \emph{Contribution to portfolio (Shapley-values)} & Calculate the approximated Shapley values indicating the contribution of each algorithm to the overall portfolios ECDF.\\ \cline{1-4}
\end{tabular}
\caption{\label{tab:summary-iohanalyzer-fixed-target}The functionalities implemented in the \emph{fixed-target results} section of \iohana. 
}
\end{table}

\subsubsection{The ``Single Function'' Subsection}
\label{sec:FT-single}

The \emph{single function} subsection offers six different types of fixed-targets results, which are grouped as follows: 
(1) data summary, (2) expected runtime, (3) probability mass function, (4) cumulative distribution, (5) algorithm parameters, and (6) statistics. These groups will be described in the following paragraphs. Note that, in the header of \iohana, there are two drop-down menus that allow the user to select the dimension and function, respectively. They are available in the sidebar when data has been loaded.
\begin{figure}[!htp]
\centering
\includegraphics[width=\textwidth, trim=0mm 0mm 0mm 0mm, clip]{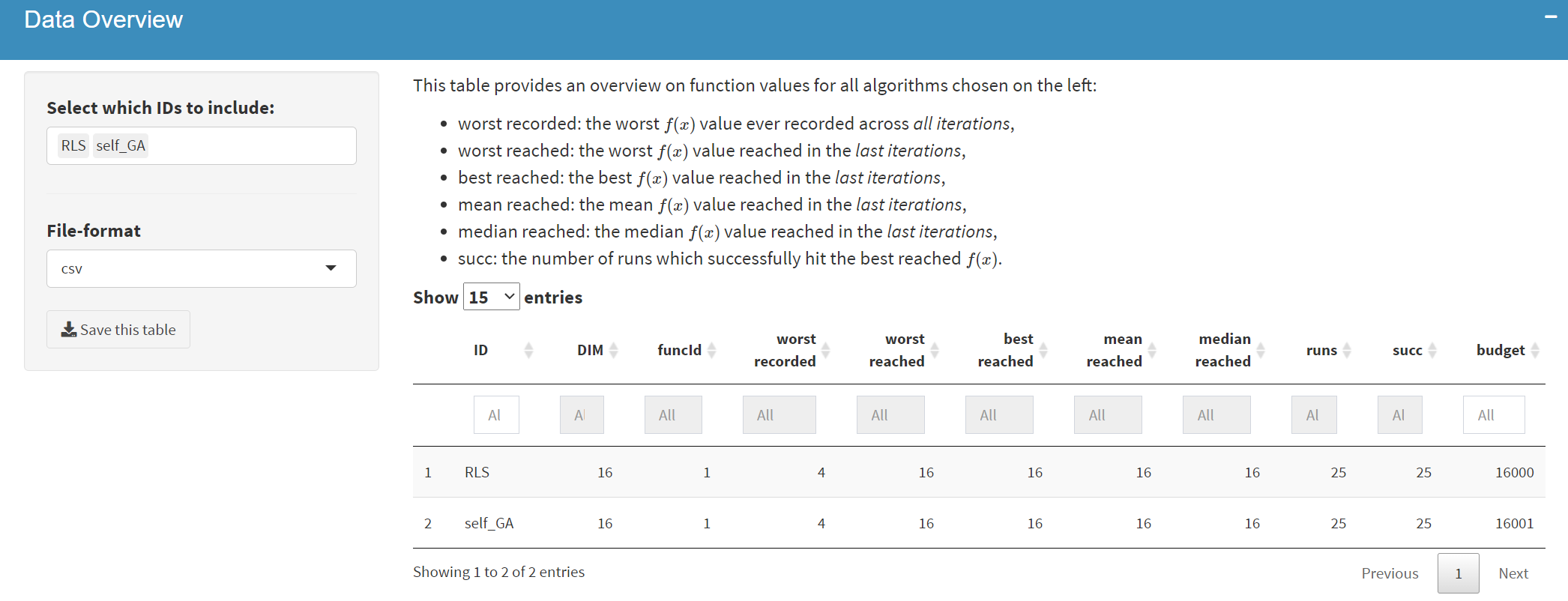}
\caption{\label{fig:RT-overview} Screenshot of the overview table of function values reached in the ``sample\_data'' data set {on function $1$ in $16$D}.
}
\end{figure}

\paragraph{Group 1: Fixed-Target Results \cascade Single Function \cascade Data Summary:} 
This group provides basic statistics on the distribution of the fixed-target running time, which are grouped in 3 different tables:
\begin{itemize}
	\item Table \texttt{Data Overview:} A screenshot of this table is given in Figure~\ref{fig:RT-overview}. It simply summarizes the range of \emph{function values} observed in the data set, with the purpose to offer its users a quick overview of the quality of the solutions that were evaluated by the algorithms. In Figure~\ref{fig:RT-overview}, we show the data overview of the ``sample\_data'' data set, where the following values are listed for each triple of function, dimension, and algorithm: 
	(1) the total number of runs, 
	(2) the worst of all function values recorded in any of the runs (``worst recorded''), 
	(3) the worst of the best function values reached in any of the runs (``worst reached''), 
	(4) the best function value reached in any of the runs (``best reached''), 
	(5) the mean (with respect to all runs) best function values (``mean reached''), 
	(6) the median (with respect to all runs) best function value (``median reached''), and 
	(7) the number of runs which successfully hit the ``best reached'' function value (``succ'').
	\item Table \texttt{Runtime Statistics at Chosen Target Values:} A screenshot of this table is given in Figure~\ref{fig:RT-summary}. The user can set the range and the granularity of the results in the box on the left. The table shows fixed-target running times for evenly spaced target values\footnote{{These target values are evenly spaced between the user-specified minimum and maximum values (whose default values are set to be the extreme values found in the data) on a linear or log scale, based on the difference in order of magnitude between the extreme values found for the specified function. This same principle is used in all similar tables and plots where both a minimum and maximum target can be chosen by the user. A notable exception are the cumulative distribution functions, where \emph{arbitrary sets} of target values can be chosen by the user}}. More precisely, 
	the table provides the success rate and the number of successful runs as defined in Eq.~\eqref{eq:success-rate}, the sample mean, median, standard deviation, the sample quantiles: $Q_{2\%}, Q_{5\%},\ldots, Q_{98\%}$, and the \emph{expected running time} (ERT) as defined in Eq.~\eqref{eq:ERT}. The user can download this table in \texttt{csv} 
	format, or as a \LaTeX{}~table.
	\item {Table \texttt{Original Runtime Samples:} This table uses the same principle as the \texttt{Runtime Statistics}, but instead displays the values for each individual run. For this table, the user can choose between a ``long'' (all sample points are arranged in a column) and a ``wide'' format (all sample points are arranged in a row).}
\end{itemize} 

\begin{figure}[!tp]
\centering
\includegraphics[width=\textwidth, trim=0mm 0mm 0mm 0mm, clip]{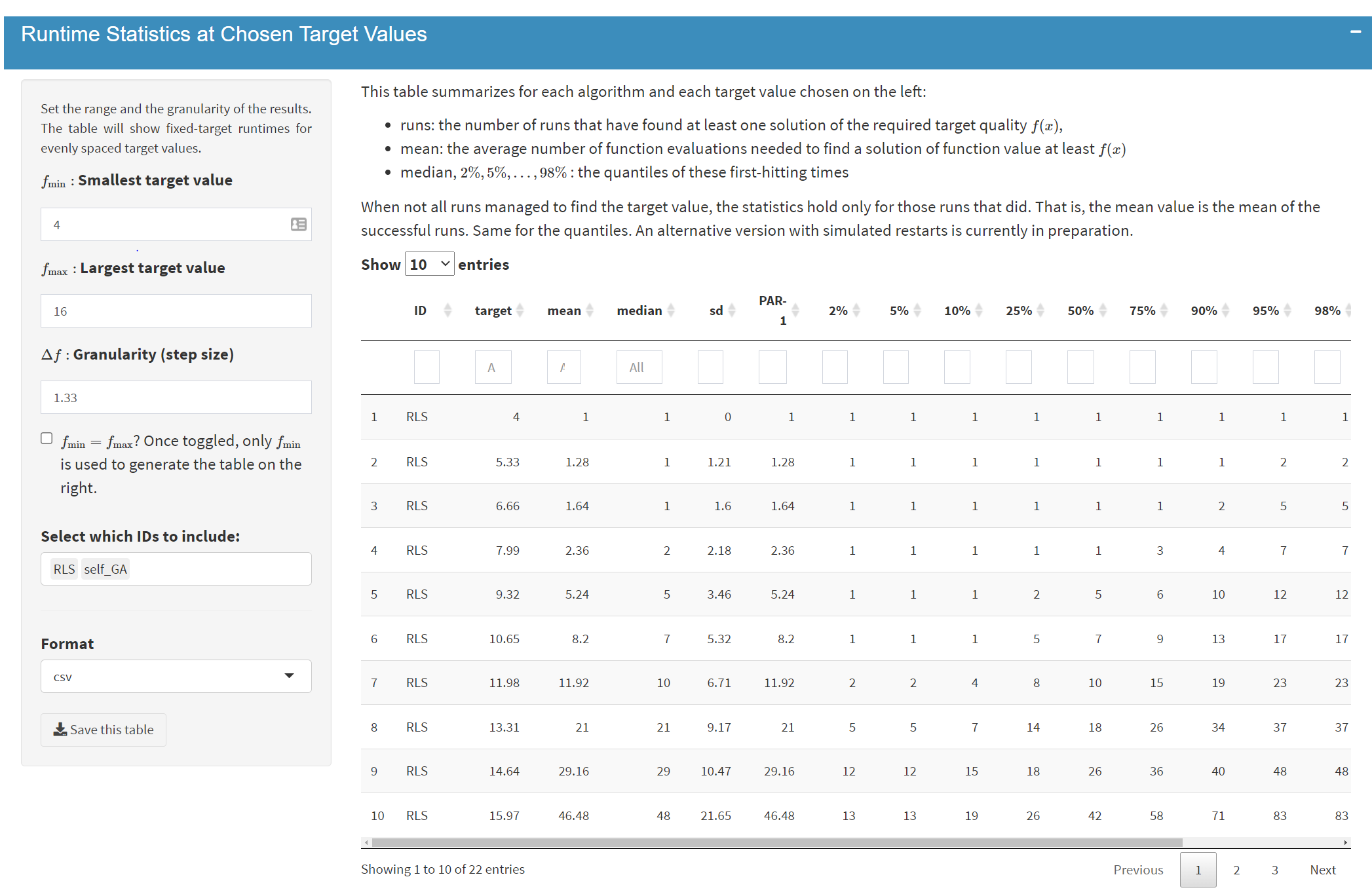}
\caption{\label{fig:RT-summary} Screenshot of the data summary table of some descriptive statistics on the running time.}
\end{figure}

\paragraph{Group 2: Fixed-Target Results \cascade Single Function \cascade Expected Runtime:} An interactive plot illustrates the fixed-target running times. An example of this plot is shown in Figure~\ref{fig:GUI-ERT-single}. The interactive plot can be adjusted in the menu on the left as shown in the figure. These options include showing/hiding mean and/or median values along with standard deviations and scaling the axes logarithmically. The user selects the algorithms to be displayed as well as the range of target values within which the curves are drawn. By default, this range is set as $[Q_{25\%}, Q_{75\%}]$ of all function values measured in the data set.
The displayed curves can be switched on and off by clicking on the legend on the bottom of the plot.

\begin{figure}[!htbp]
\centering
\includegraphics[width=\textwidth, trim=0mm 0mm 0mm 0mm, clip]{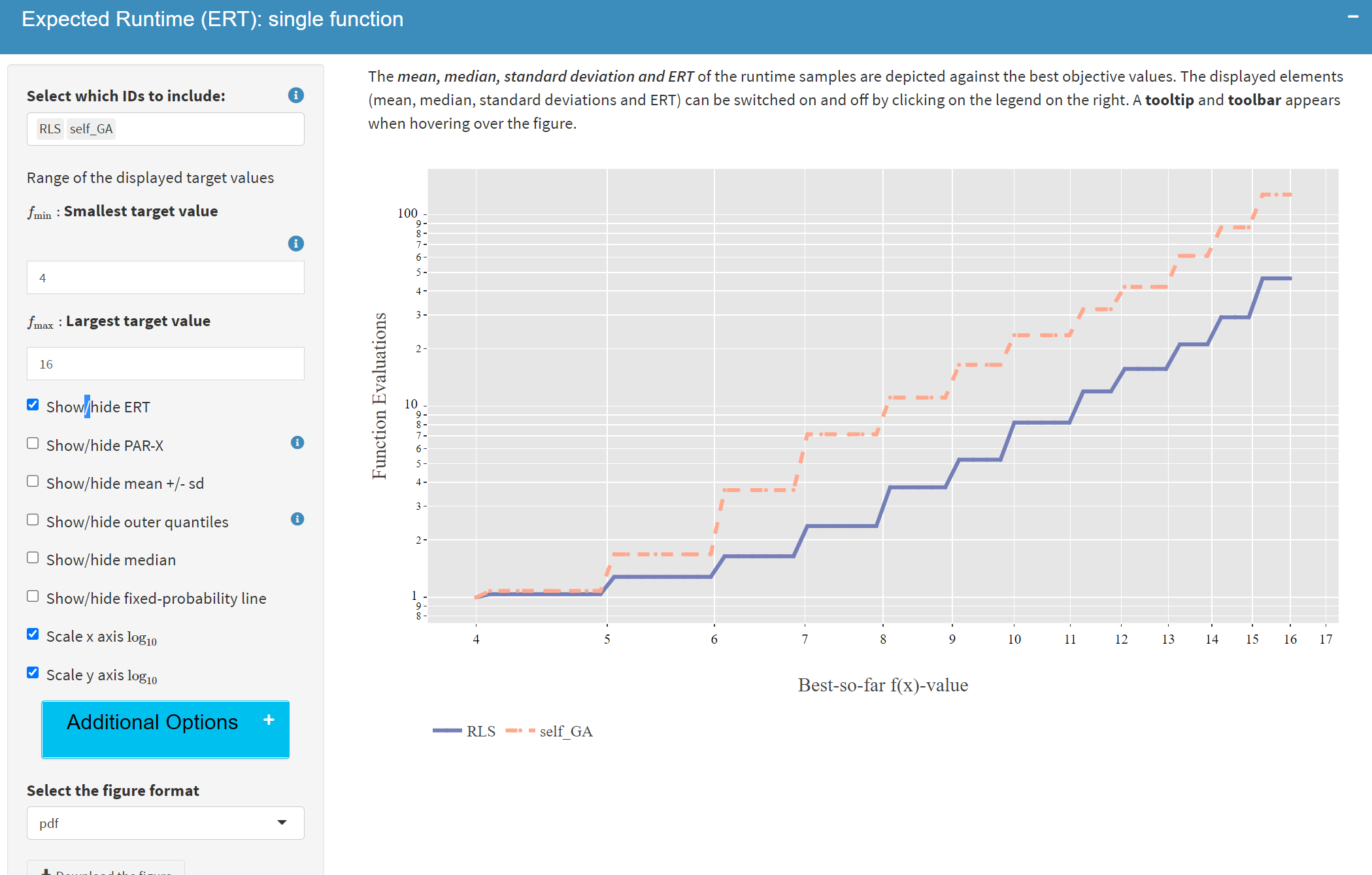}
\caption{\label{fig:GUI-ERT-single} Screenshot of the expected running time plot. }
\end{figure}

\paragraph{Group 3: Fixed-Target Results \cascade Single Function \cascade Probability Mass Function:} 
For a selected target value $v$, the histogram of the running time, as displayed in Figure~\ref{fig:GUI-RT-hist}, shows the number of runs $i$ for which the running time falls into a given interval $[t,t+\Delta t)$, namely $\sum_{i=1}^r \mathds{1}(t \le T(A,f,d,v,i) < t + \Delta t)$. The bin size $\Delta t$ is determined according to the \emph{Freedman-Diaconis} rule~\cite{Freedman1981}, which is based on the interquartile range of the sample $\{T(A,f,d,v,i)\}_{i=1}^r$.
The user has two options: 1) an \emph{overlayed display}, where all algorithms are displayed in the same plot, or 2) a \emph{separated one}, where each algorithm is displayed in an individual sub-plot, as shown in Figure~\ref{fig:GUI-RT-hist}.

In addition to the histogram, the probability mass function (Figure~\ref{fig:GUI-RT-pmf}) might be helpful to get a finer look at the shape of the empirical distribution of $T$. The user can switch on/off the illustration of all sample points (depicted as dots), or only the empirical probability mass function itself. It is important to point out that the probability mass function is estimated in a ``continuous'' manner, where running time samples are considered as $\mathbb{R}$-valued and then the \emph{Kernel Density Estimation} (KDE) method is taken to estimated the function.\footnote{Strictly speaking, this method gives imprecise estimations when there are many duplicated values, which can be quite likely in discrete optimization (such as in our examples). Improvements are planned for the future version.} 
\begin{figure}[!htbp]
\centering
\includegraphics[width=.85\textwidth, trim=0mm 0mm 0mm 0mm, clip]{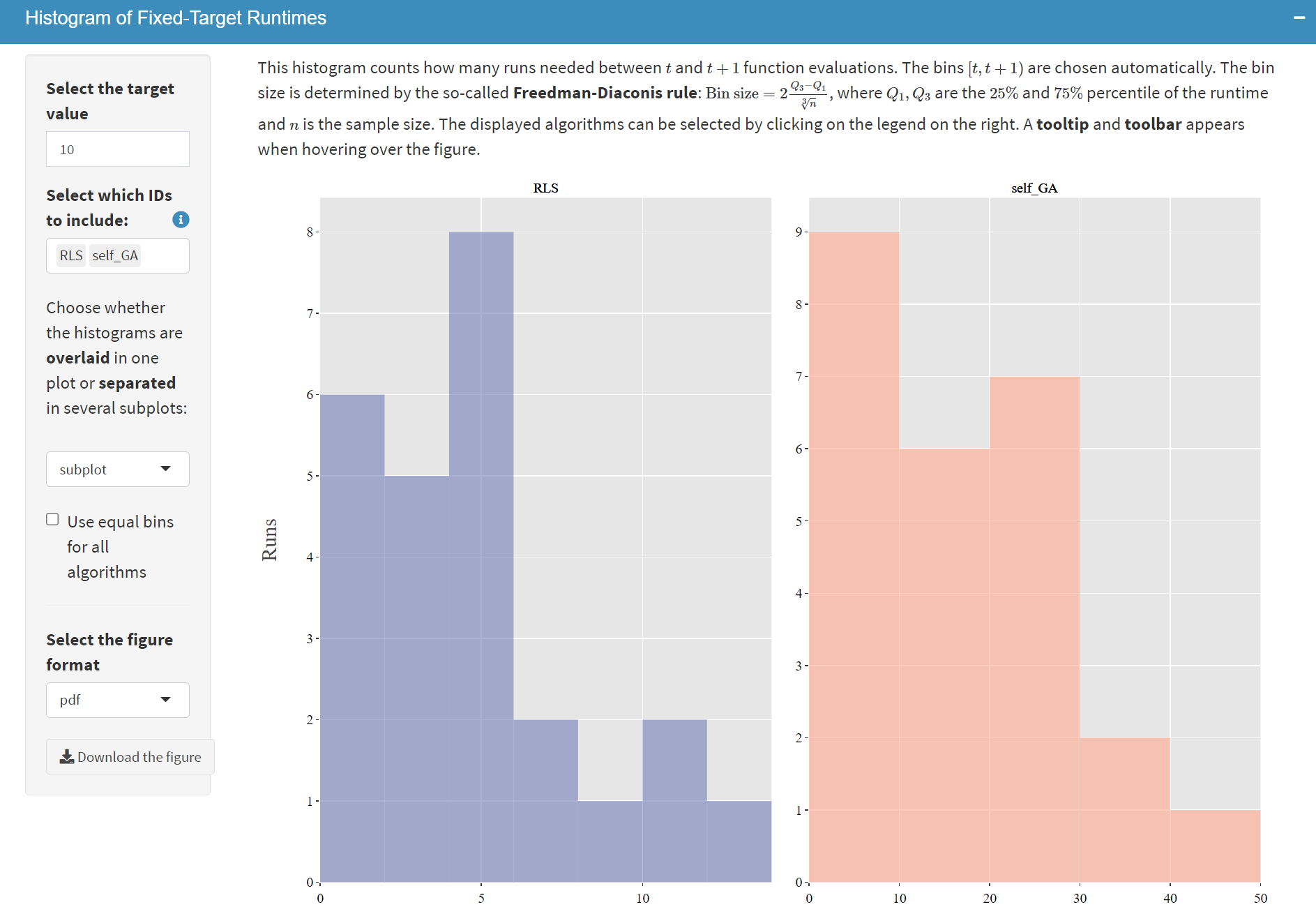}
\caption{\label{fig:GUI-RT-hist} Screenshot of the histogram of running time (given a fixed target of $10$ on Function 1 in dimension 16).}
\end{figure} 
\begin{figure}[!htbp]
\centering
\includegraphics[width=.85\textwidth, trim=0mm 0mm 0mm 0mm, clip]{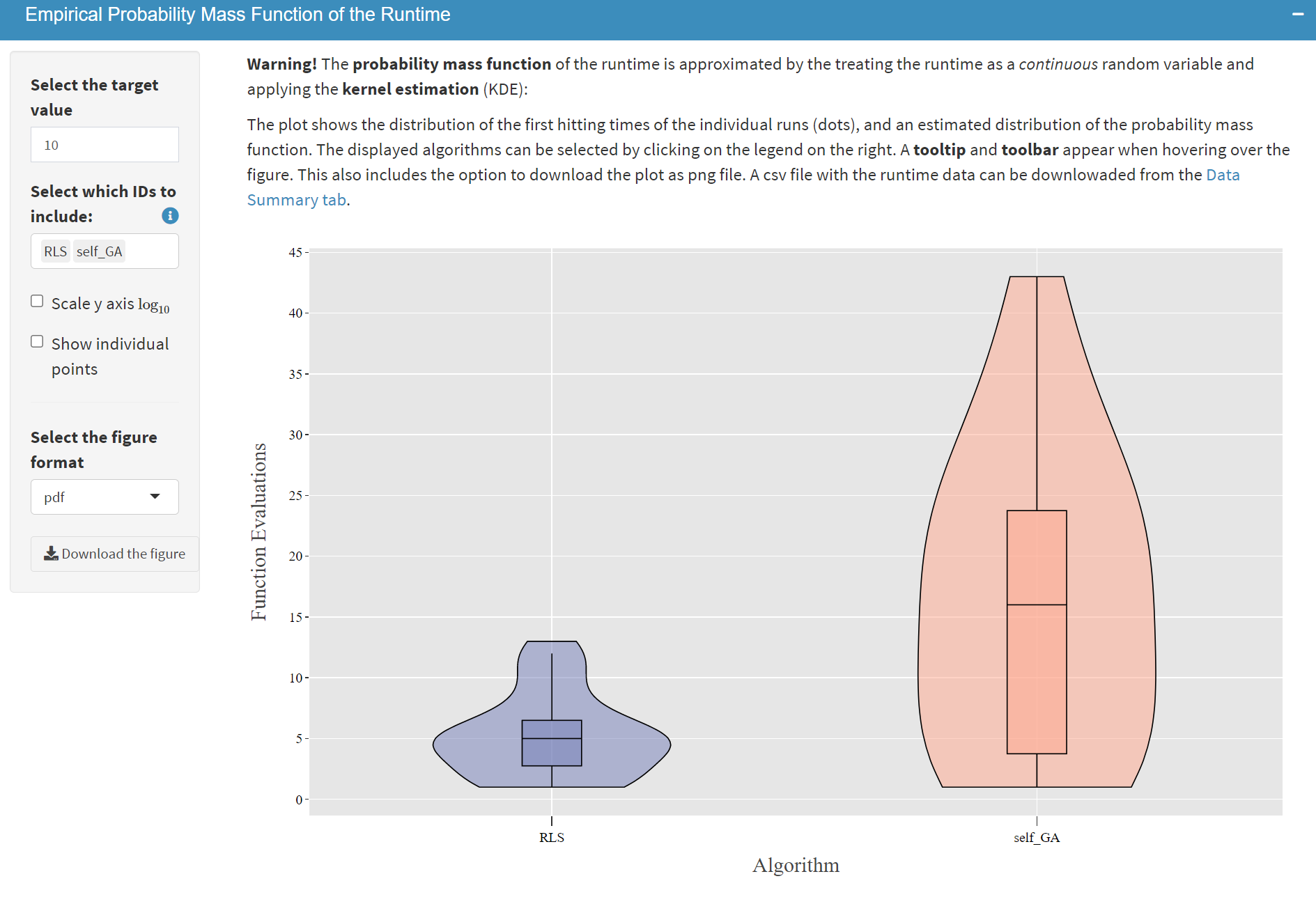}
\caption{\label{fig:GUI-RT-pmf} Screenshot of the empirical probability mass function of running time for target value $10$ on Function~1 in dimension~16).}
\end{figure}

\paragraph{Group 4: Fixed-Target Results \cascade Single Function \cascade Cumulative Distribution} 
The empirical cumulative distribution function (ECDF) of the running time is computed for target values specified by the user. In addition to calculating ECDFs for a single target value, it is recommended to aggregate ECDFs over multiple targets, to obtain an overall performance for solving different targets. For the default target values, the tool takes $10$ evenly spaced values in $[Q_{25\%},Q_{75\%}]$ of all measured function values in a data set. Such a functionality is exemplified in Figure~\ref{fig:GUI-ECDF-target}: a set of evenly spaced target values can be generated by specifying the range and step-size of the target value. 

In this example, with the following setup, $f_{\min}=4$, $f_{\max}=16$, and $\Delta f=1.33$, the target value sequence, $4, 5.33, 6.66,\ldots, 16$ is used to calculate the ECDF. These values are shown in the bar on the top of the plot, as can be seen in Figure~\ref{fig:GUI-ECDF-target}. 
In the same figure it can be seen for algorithm RLS (blue curve) that within a budget of 24 function evaluations, around $76\%$ of (target, run) pairs have been successful. For algorithm \oplga (purple curve) this value is only $53\%$.

\begin{figure}[!htbp]
\centering
\includegraphics[width=\textwidth, trim=0mm 0mm 0mm 0mm, clip]{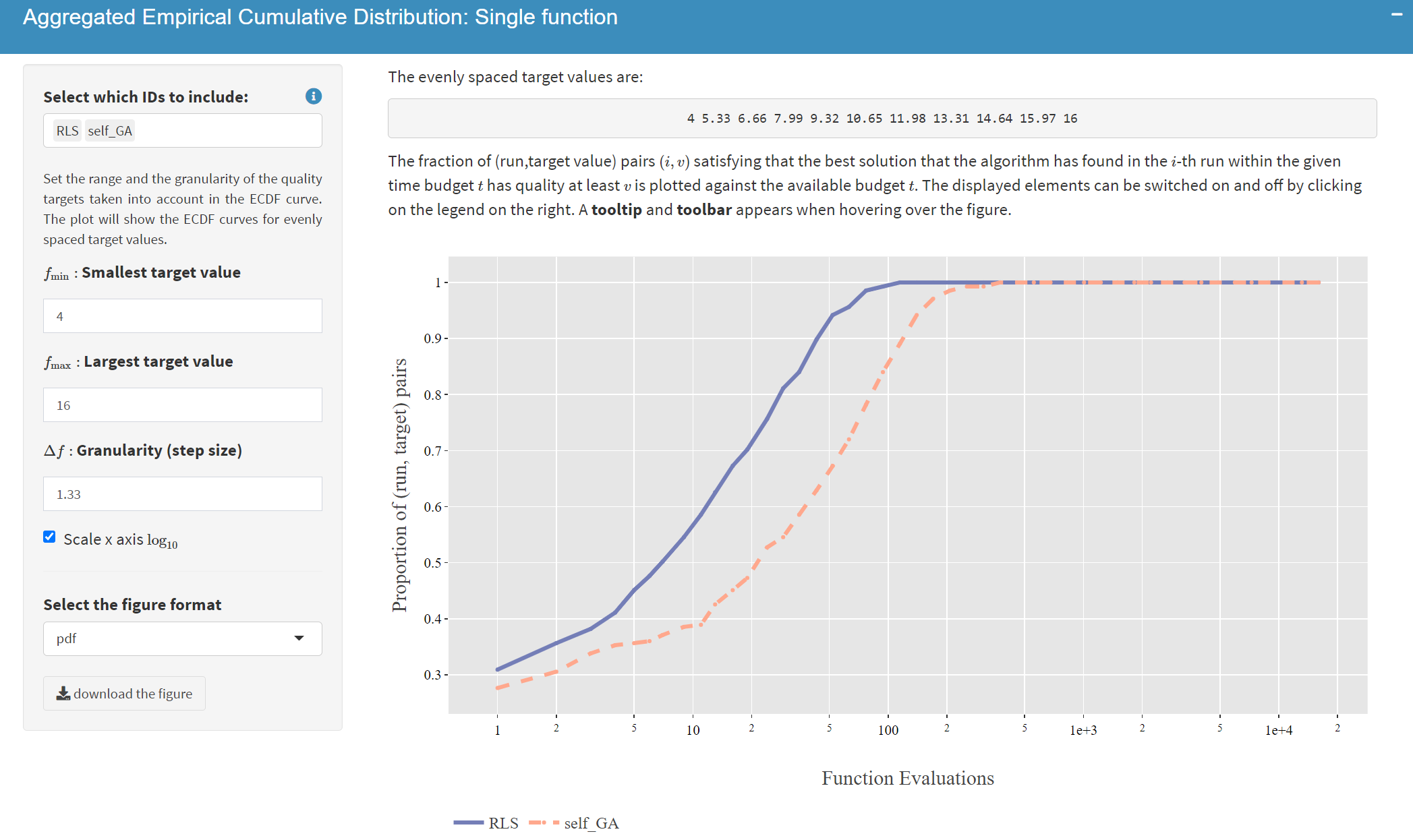}
\caption{\label{fig:GUI-ECDF-target}  Screenshot of aggregated ECDF curve over multiple targets.}
\end{figure}

\paragraph{Group 5: Fixed-Target Results \cascade Single Function \cascade Algorithm Parameters} 
One of the key motivations to build \iohpro was the ability to analyze, in detail, the evolution of control parameters which are adjusted during the search. Such dynamic parameters can be found in most state-of-the-art heuristics. While in numerical optimization a non-static choice of the search radius, for example, is needed to eventually converge to a local optimum, dynamic parameters are also more and more common in discrete  and mixed-integer optimization heuristics~\cite{KarafotiasHE15,DoerrD20chapter}. In the fifth group of fixed-target results for a single function, the evolution of the parameters is linked to the quality of the best-so-far solutions that have been evaluated. In the experimentation (i.e., data generation) phase, the user selects which parameters are logged along with the evaluated function values. These values are then automatically detected by \iohana and can be chosen in this group for analysis.

As with the interactive plots on expected running time, the user can choose the range of targets, which parameters and algorithms to plot, and the scale (either logarithmic or linear) of $x$- and $y$-axis. We omit the example for parameters as the GUI is similar to the one in Figure~\ref{fig:GUI-ERT-single}. As with  ``Fixed-Target Results \cascade Single Function \cascade Data Summary'', this subsection also provides for each parameter tables of descriptive statistics (sample mean, median, standard deviation, and some quantiles) as well as the original parameter values.
\begin{figure}[!ht]
\centering
\includegraphics[width=\textwidth, trim=0mm 00mm 0mm 0mm, clip]{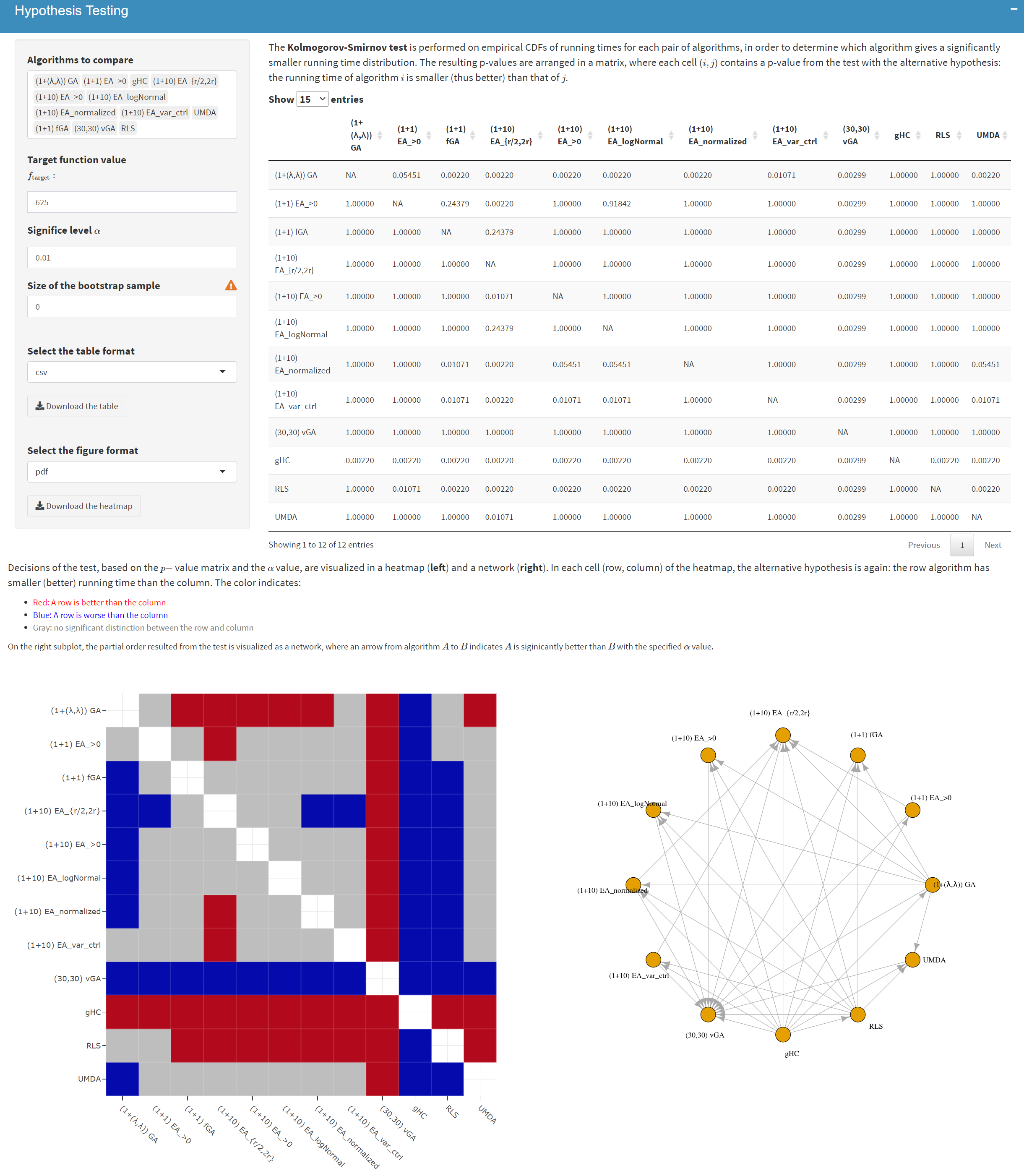}
\caption{\label{fig:GUI-Statistics} Screenshot of the multiple testing procedure applied on all {$12$} reference algorithms on function $f1$ and dimension {$625$}. {The table shows the p-values resulting from the pairwise KS-test between each pair of algorithms. Then, based on the $\alpha=0.01$, the resulting hypothesis-rejections are shown in both the matrix-plot and the network.}
}
\end{figure}

\paragraph{Group 5: Fixed-Target Results \cascade Single Function \cascade Statistics} 
To address the robustness of empirical comparisons, the samples from all algorithm must undergo a proper statistical test procedure~\cite{hollander2013nonparametric}. In \iohana, a standard multiple testing procedure is implemented to compare the fixed-target running time for each pair of algorithms on a single function, for which the well-known Kolmogorov-Smirnov test is applied to the ECDFs of running times. Moreover, the Bonferroni procedure is used to correct the $p$-value in multiple testing. To demonstrate this functionality, we show, in Figure~\ref{fig:GUI-Statistics}, the testing outcome of a data set on $12$ reference algorithms.\footnote{This data set is available at \url{https://github.com/IOHprofiler/IOHdata/blob/master/iohprofiler/2019gecco-ins1-11run.rds} It can be loaded to the web-based GUI by selecting the PBO data set in the ``upload data'' section. The data set comprises the results of the experimental study described in~\cite{DoerrYHWSB20}.} on the PBO problem set from~\cite{DoerrYHWSB20}, instead of the exemplary two-algorithm data set used previously. Here, the test is conducted across all $12$ algorithms on function $f1$ and dimension $64$ with a confidence level of $0.01$.
The result of this procedure is illustrated by a table of pairwise $p$-values, a color matrix of the statistical decision, and a graph depicting the partial order induced by the test (i.e., an arrow pointing from Algorithm~1 to Algorithm~2 is to be read as Algorithm~1 dominating Algorithm~2 with statistical significance. As with all tables and figures in \iohana, these can be downloaded in several formats, including \texttt{*.tex} and \texttt{*.csv} for tables and \texttt{*.pdf} and \texttt{*.eps} for figures.

\subsubsection{The ``Multiple Functions'' Subsection}
\label{sec:FT-multi}
This subsection contains three groups of fixed-target results for multiple functions: (1) expected runtime comparison across all functions for one dimension, (2) aggregated Empirical Cumulative Distribution over all functions, and (3) Glicko2-based ranking.

\paragraph{Group 1: Fixed-Target Results \cascade Multiple Functions \cascade Expected Runtime:} In this group, the tool depicts the ERT values against multiple functions as a radar-plot, as shown in Figure~\ref{fig:GUI-ERT-MULTI}. For each function, the target value used for calculating the ERT is determined by default as follows: firstly, for each algorithm, we obtained the 2\% percentile of the best function values reached in multiple runs. Secondly, we took the largest value among all such 2\% percentiles as the target value on this function. 
In this radar-plot, we revert the axis such that the bigger ERT values are further away from the center of the circle compared to smaller ones, indicating that better algorithms will cover a larger area.
\begin{figure}[!ht]
\centering
\includegraphics[width=\textwidth, trim=0mm 0mm 0mm 0mm, clip]{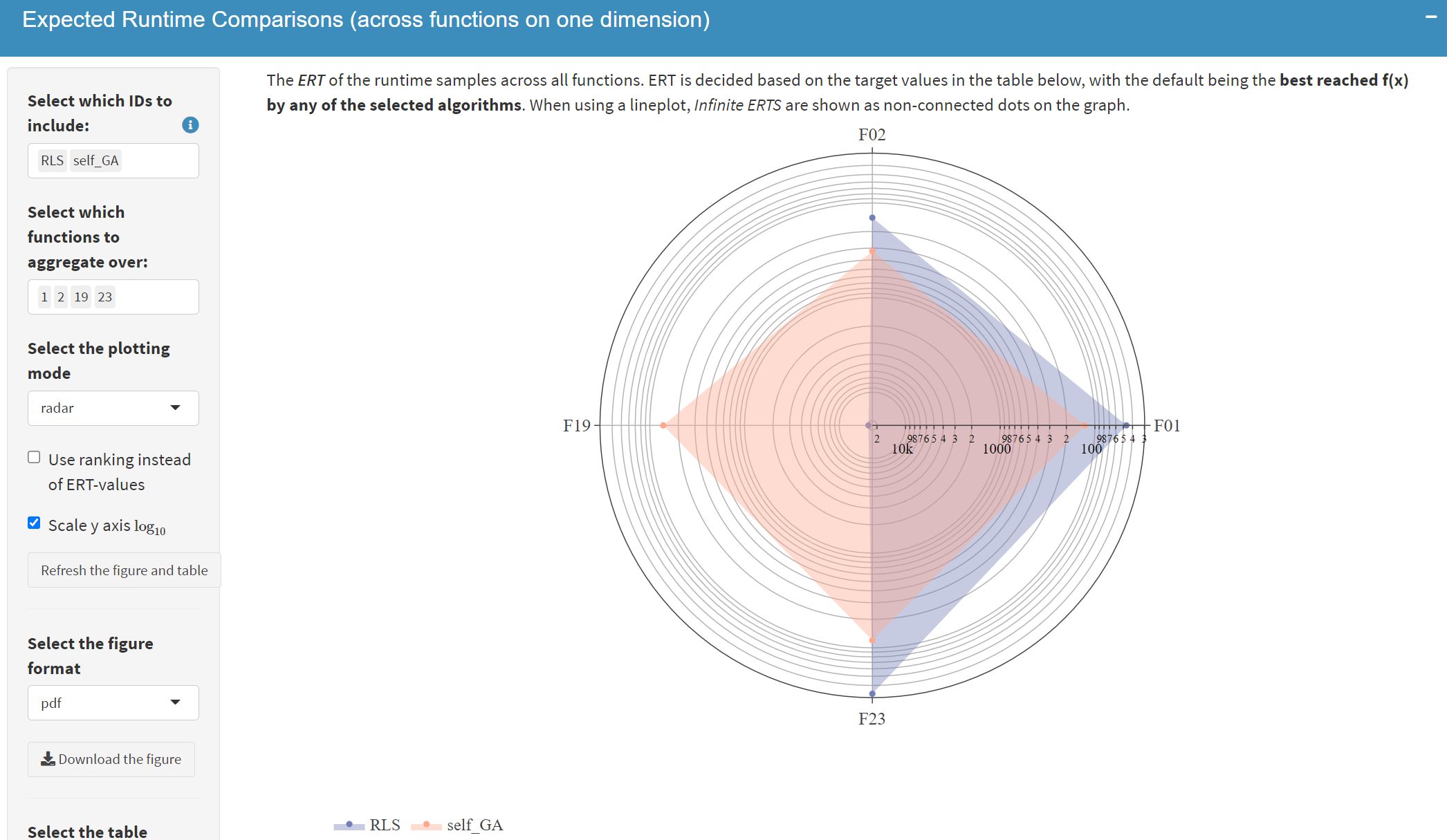}
\caption{\label{fig:GUI-ERT-MULTI} Screenshot of ERT of RLS and the \oplga on four PBO problems, where the ERT values are shown for each selected problem in a radar plot with inverted axis (values are decreasing when moving away from the center). Loosely speaking, an algorithm with a larger span on the plot is considered better, e.g., RLS dominates \oplga on problem F01, F02, and F23 while the latter is superior on F19.}
\end{figure}

\paragraph{Group 2: Fixed-Target Results \cascade Multiple Functions \cascade Cumulative Distribution:}
In this group, ECDFs of running times are aggregated across multiple functions, as defined in Eq.~\eqref{eq:ECDF-agg-fct}. This functionality is illustrated in Figure~\ref{fig:GUI-ECDF-fct}: a table of pre-calculated target values are provided for each function (all test functions are included by default). This table of targets can easily be edited directly in the GUI, or by a downloading-editing-uploading procedure (which should, of course, not change the format of the tables, just the values).
\begin{figure}[!htbp]
\centering
\includegraphics[width=\textwidth, trim=0 0 0 0, clip]{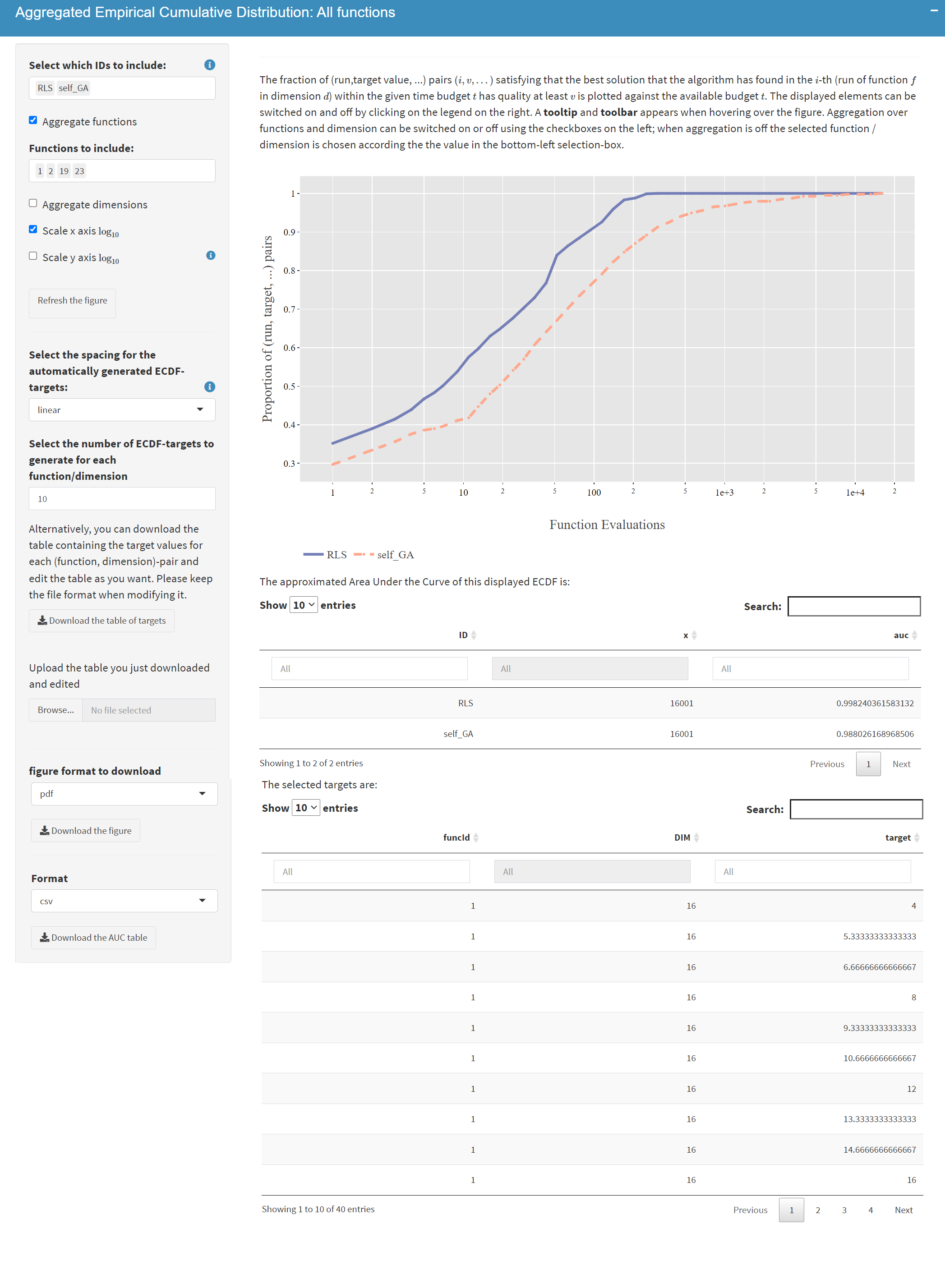}
\caption{\label{fig:GUI-ECDF-fct} Screenshot of aggregated ECDF curve across multiple functions and targets.}
\end{figure}
{Note that for these ECDF-figures, the corresponding Area Under the Curve (AUC) can also be calculated to get a single value for each algorithm. These AUC-tables are available in the same tab as the ECDF plot.}

\paragraph{Group 3: Fixed-Target Results \cascade Multiple Functions \cascade Ranking:}
This group provides a ranking functionality to compare algorithms across multiple functions and dimensions, in which we employ the Glicko-2 rating system~\cite{glickman2012example} (commonly used in chess games) to rank the algorithms, based on multiple simulated games between them ($25$ by default). In each game, for every function and dimension, the winner of each pair of algorithms is determined by sampling from the running time values (given a target value) uniformly at random and checking which random sample is better. An overall rating is computed from those games' outcome, which is then used to rank the algorithms. {Algorithms with a better rank win more rounds than those with a poor rank, indicating that when sampling a runtime on an arbitrary function, these algorithms tend to perform better. This way of ranking allows us to aggregate performance over an arbitrary number of functions and targets, while inherently managing uncertainty of the performance metrics by doing the repeated rounds and comparing individual values for each 'game'.}

\subsection{The ``Fixed-Budget Results'' Section}
\label{sec:FB}

The fixed-budget section offers performance analysis for the quality of the best solution that the algorithms could identify within a given budget of function evaluations. The results are similar to those presented in the fixed-target section (Section~\ref{subsec:fixed-target-results}) except that subsection ``Fixed-Budget Results \cascade Multiple Functions \cascade Cumulative Distribution'' is still under development and hence not yet available at the time of writing. Table~\ref{tab:summary-iohanalyzer-fixed-budget} summarizes the main functionalities. 

\begin{table}[!htbp]
\scriptsize
\centering
\renewcommand{\arraystretch}{1.3}
\begin{tabular}{|M{.6cm}|m{1.7cm}|m{3cm}|m{6cm}|}
\hline
Section & Group &  Functionality  & Description \\ \hline\hline
\multirow{22}{*}{\rotatebox[origin=r]{90}{Single Function}} & \multirow[c]{5}{1.5cm}{Data Summary}  &\emph{Data Overview} & The minimum and maximum of running times for selected algorithms. \\ \cline{3-4}
& & \emph{Target Value Statistics} & The mean, median, quantiles of the function value at a sequence of budgets controlled by $B_{\text{min}},B_{\text{max}}$ and $\Delta B$.\\ \cline{3-4}
& & \emph{Target Value Samples} & The function value samples at an evenly spaced sequence of budgets controlled by $B_{\text{min}},B_{\text{max}}$ and $\Delta B$. \\\cline{2-4}
& \parbox[c][.8cm]{1.3cm}{Expected Target Value} & \parbox[c][.8cm]{3cm}{\emph{Expected Target Value: single function}} & \parbox[c][.8cm]{6cm}{The progression of expected function values over budgets, whose range is controlled by the user.} \\\cline{2-4}
& \multirow[c]{3}{1.5cm}{Probability Density Function} & \emph{Histogram} & The histogram of the function value a user-chosen budget. \\\cline{3-4}
& & \emph{Probability Density Function} & The probability density function (using the Kernel Density Estimation) of the function value at a user-chosen budget. \\\cline{2-4}
& \multirow[c]{4}{1.5cm}{Cumulative Distribution}  & \emph{ECDF: single budget}& On \textbf{one} function, the ECDF of the function value at \textbf{one} budget specified by 
the user. \\ \cline{3-4}
& & \emph{ECDF: single function}& On \textbf{one} function, ECDFs aggregated over \textbf{multiple} budgets. \\ \cline{3-4}
& & \emph{Area Under the ECDF} & On \textbf{one} functions, the area under ECDFs of function values that are aggregated over \textbf{multiple} budgets. \\ \cline{2-4}
& \multirow[c]{5}{1.5cm}{Algorithm Parameters}  & \emph{Expected Parameter Value} & The progression of expected value of parameters over the \textbf{budget}, whose range is controlled by the user. \\ \cline{3-4}
& & \emph{Parameter Statistics} & The mean, median, quantiles of recorded parameters at an evenly spaced sequence of budgets controlled by $B_{\text{min}},B_{\text{max}}$ and $\Delta B$. \\ \cline{3-4}
& & \emph{Parameter Sample} & The sample of recorded parameters at an evenly spaced sequence of budgets controlled by $B_{\text{min}},B_{\text{max}}$ and $\Delta B$. \\ \cline{2-4}  
& Statistics & \emph{Hypothesis Testing} & The two-sample Kolmogorov-Smirnov test applied on the running time at a target value for each pair of algorithms. A partial order among algorithms is obtained from the test \\ \cline{1-4}
\multirow{15}{*}{\rotatebox[origin=r]{90}{Multiple Functions}}& \multirow[c]{2}{1.5cm}{{Data Summary}} & \emph{Multi-Function Statistics
} & Descriptive statistics for all functions at a single target value. \\ \cline{3-4}
& & \emph{Multi-Function Hitting Times
} & Raw hitting times for all functions at a single target value. \\ \cline{2-4}
& \multirow[c]{3.6}{1.5cm}{Expected Target Value} & \emph{Expected Target Value: all functions} & The same as above expect that the expected function values are grouped by functions and the range of budgets are automatically determined. \\ \cline{3-4}
& & \emph{Expected Target Value: Comparison} & The expected function value at the largest budget found on each function is plotted against the function ID for each algorithm.\\\cline{2-4}
& \multirow[c]{4.2}{2.5cm}{{Deep Statistics}} & \emph{Ranking per Function} & Per-function statistical ranking procedure from the Deep Statistical Comparison Tool (DSCTool)~\cite{EftimovPK20}.\\ \cline{3-4}
& & \emph{Omnibus Test} & Use the results of the per-function ranking to perform an omnibus test using DSC. \\ \cline{3-4}
& & \emph{Posthoc comparison} & Use the results of the omnibus test to perform the post-hoc comparison. \\ \cline{2-4}
& Ranking & \emph{Glicko2-based ranking} & For each pair of algorithms, a function value at a given budget is randomly chosen from all sample points in each round of the comparison. The glicko2-rating is used to determine the overall ranking from all comparisons. \\ \cline{1-4}
\end{tabular}
\caption{\label{tab:summary-iohanalyzer-fixed-budget}The functionalities implemented in the \emph{fixed-budget results} of \iohana. 
}
\end{table}

\subsection{The ``Position Information'' Section}
\label{sec:position}
Within this section, the user can visualize the final search points in their decision space in a parallel coordinate plot. {In version 0.1.6.1, this functionality is only supported for data generated by the \pkg{SOS}-framework~\cite{caraffini2020sos}. A processed dataset in this format is available on IOHdata.\footnote{\url{https://github.com/IOHprofiler/IOHdata/tree/master/SOS}} This dataset contains a DE-variant which was generated for the analysis of Structural Bias in DE~\cite{Anisotroy_structural_bias}, which can be confirmed visually using the parallel coordinate plot functionality.}

Development on extending this position-based functionality to other data sources and more types of analysis is in progress.

\subsection{Command-Line interface}
In addition to the web-based graphical interface, we provide a command-line interface (CLI) via a feature-rich \proglang{R}-package, which allows for more fine-grained control of various types of analysis and visualization described in this section. {All functionality discussed in this paper can also be accessed through this CLI. A demonstration of the key aspects of this CLI on an example data set is available on the wiki page \url{https://iohprofiler.github.io/IOHanalyzer/R/}.} 



\section{Discussion and Outlook} \label{sec:summary}\label{sec:outlook}

We have presented \iohana, a highly versatile environment for evaluating the performance data of iterative optimization heuristics. \iohana\ -- and, more generally, the whole \iohpro project -- are under continuous development. Extensions planned for the near future comprise, most notably, an \textbf{increased compatibility with the following benchmarking environments and platforms}:
\begin{itemize}
	\item \textbf{General-purpose benchmarking platforms.} 
	As mentioned, \iohana has already been extended to visualize data sets generated with Facebook's Nevergrad platform~\cite{nevergrad}. We are now working on various other interfaces, which will allow Nevergrad users to use the logging functionalities of \iohpro and to access the problems made available in \iohpro. Likewise, we are working towards an interface that allows users of \iohpro to more easily access the benchmark problems of Nevergrad. 
    \item \textbf{Modular algorithm frameworks and automated configuration tools.} The modular algorithm framework ParadisEO~\cite{ParadisEO2004} and the modular CMA-ES framework proposed in~\cite{modular-CMAES} have already been integrated into IOHprofiler. An integration of other modular algorithm frameworks such as those 
    presented in~\cite{jMetal15,ECJ1,OpenBeagle,DEAP}, 
    together with automated algorithm configuration tools such as irace~\cite{LopezDCBS16}, SMAC~\cite{HutterHL11}, hyperband~\cite{LiJDRT17}, and MIP-EGO~\cite{WangSEB17}, will pave the way to a more generic research environment for automated configuration of optimization algorithms. 
    For supervised learning approaches, we shall interface \iohpro and feature-extraction techniques such as those collected in the R package \emph{flacco}~\cite{flacco}.
    \item \textbf{Collections and generators of benchmark problems.} As we are doing for the Nevergrad platform, we are working on easier interfaces with other collections of benchmark problems as well as with generators of these. Already implemented are the 23 discrete problems described in the PBO suite from~\cite{DoerrYHWSB20}, a (slight variation of the)  W-model~\cite{Weise:2018} (see~\url{https://iohprofiler.github.io/} for details of our implementation), and the 24 numeric optimization problems from the BBOB suite~\cite{HansenFRA09} of the COCO platform~\cite{hansen2016coco}. {Extensions to other problem types, such as multi-objective or noisy optimization, are also being considered.}
	\item \textbf{Other statistical evaluation techniques.} Several interfaces of \iohana with tools aimed at visualizing or analyzing the performance data are currently under consideration. For example, an integration of the software to efficiently compute \emph{empirical attainment functions} provided by~\cite{eaf} could help to visualize the time-quality-robustness trade-off of IOHs. 
	
	Building on the initial study~\cite{CalvoSCD0BL19} we are considering the integration of the rank-based Bayesian inference statistics, which were introduced to the evolutionary computation community via~\cite{Calvo:2018:BIA:3205651.3205658}. Other advanced statistical procedures may also be added, e.g., the \emph{Deep Statistical Comparison tool} (DSCtool) suggested in~\cite{EFTIMOV2019255}. 
	\item\textbf{Performance aggregation {and anytime performance measures}.} 
	Finally, we are also implementing different ways to aggregate performances over multiple test problems. In this respect we are, among others, looking into so-called performance profiles~\cite{more2009benchmarking}, which is the empirical cumulative distribution of normalized performance values across problems. {Related to this, we observe an increasing interest in measuring and/or optimizing for anytime performance metrics~\cite{JesusLDP20,BossekKT20}. We are carefully observing this development and are considering different ways to extend the statistics of \iohana with other suggested anytime performance measures.}  
    
\end{itemize}
 

\textbf{Computational details}
{The results in this paper were obtained using
\proglang{R}~4.1.1 and version 0.1.6.1 of \iohana with the following packages, \pkg{Rcpp}~1.0.7, \pkg{shiny}~1.6.0 and \pkg{plotly}~4.9.4.1. For the \proglang{C/C++} compiler, gcc version~10.3.0 is used (for \pkg{Rcpp}).}

\subsection*{Acknowledgments}
We thank 
Arina Buzdalova, Maxim Buzdalov, Rapha\"el Cosson, Johann Dr\'eo, Tome Eftimov, Pietro S. Oliveto, Ofer M. Shir, Markus Wagner, and Thomas Weise for various suggestions that have helped us improve \iohana and the presentation of this tool. We also thank the COCO team, in particular Anne Auger, Dimo Brockhoff, and Nikolaus Hansen, as well as the Nevergrad team, Jeremy Rapin and Olivier Teyaud, for help with their platforms. We also thank the anonymous reviewers of ACM TELO whose comments have helped us to improve the presentation and reproducibility of this work.  

Our work has been financially supported by the Paris Ile-de-France region and by a public grant as part of the Investissement d'avenir project, reference ANR-11-LABX-0056-LMH, LabEx LMH, in a joint call with the Gaspard Monge Program for optimization, operations research, and their interactions with data sciences. 
Furong Ye acknowledges financial support from the China Scholarship Council, CSC No. 201706310143. Parts of our work have been inspired by working group 3 of COST Action CA15140 `Improving Applicability of Nature-Inspired Optimisation by Joining Theory and Practice (ImAppNIO)' supported by the European Cooperation in Science and Technology.
 


\newcommand{\etalchar}[1]{$^{#1}$}

\appendix
\section{Supported Data Format}\label{subsec:data-format}
\iohana aims to be as flexible as possible, and to achieve this, it supports data from many different sources. This means that data can be presented in many different formats. At the time of writing, the list of supported formats is as follows:
\begin{itemize}
	\item \iohpro data format, which is motivated by and modified from the \pkg{COCO} data format.
	\item \pkg{COCO} data format as defined in~\cite{hansen:inria-00362649}. 
	\item The \pkg{Nevergrad} format from~\cite{nevergrad}. 
	\item The \pkg{SOS} format from~\cite{caraffini2020sos}.
	\item A ``two-column'' format is a simplified version of the \iohpro format. We describe this format in the upcoming paragraph on ``Raw-data''. 
\end{itemize}
When loading the data in the programming interface (and in  the graphical user interface as well), it is not necessary to specify its format as \iohana attempts to detect this automatically. For most data formats,\footnote{The \iohpro, \pkg{COCO} and the \emph{two-column} formats have the same basic structure, while Nevergrad uses pure csv files instead, and will thus not be discussed in this section.} data files are organized in the same manner within the file system. The structure of data files is as follows:
\dirtree{%
.1 {.}/.
.2 IOHprofiler\_f1{.}info.
.2 data\_f1.
.3 IOHprofiler\_f1\_DIM64{.}dat.
.3 IOHprofiler\_f1\_DIM64{.}cdat.
.3 IOHprofiler\_f1\_DIM100{.}dat.
.3 IOHprofiler\_f1\_DIM100{.}cdat.
.3 $\ldots$.
.2 IOHprofiler\_f2{.}info.
.2 data\_f2.
.3 IOHprofiler\_f2\_DIM64{.}dat.
.3 IOHprofiler\_f2\_DIM64{.}cdat.
.3 IOHprofiler\_f2\_DIM100{.}dat.
.3 IOHprofiler\_f2\_DIM100{.}cdat.
.3 $\ldots$.
.2 IOHprofiler\_f3{.}info.
.2 $\ldots$.
}

Generally, in the folder (e.g., \verb|./| here) that contains the data set, the following files are mandatory for \iohana: 
\begin{itemize}
	\item \emph{Meta-data} (\verb|.info|) files summarize the algorithmic performance for each problem instance, with the following naming convention:
		\verb|IOHprofiler_f1.info| for problem $f1$. Note that one meta-data file can consist of several dimensions. Please see the details below.
	\item \emph{Raw-data} (\verb|.dat|, \verb|.cdat| etc) are \texttt{csv}-like files that contain performance information indexed by the running time. Raw-data files are named in a similar manner as the meta-data files, for example, \verb|IOHprofiler_f1_DIM100.dat| for problem $f1$ and dimension $100$. Raw-data files are organized in sub-folders for each problem. It is important to note that those three data formats only differ in the structure of the raw-data files.
\end{itemize}

\paragraph{Meta-data}
When benchmarking, it is common to specify a number of different dimensions, functions and instances, resulting in a quite large number of data files (e.g., \verb|*.dat| files). It would make the data organization more structured if some meta data are provided. Here, the meta data are implemented in a format that is very similar to that in the well-known \texttt{COCO} environment. The meta data are indicated with suffix \verb|.info|. A small example is provided as follows:

\begin{CodeChunk}
\small
\begin{CodeInput}
suite = 'PBO', funcId = 19, DIM = 16, algId = 'self_GA'
%
data_f19/IOHprofiler_f19_DIM16.dat, 1:16001|3.20000e+001, 1:16001|3.20000e+001, 
1:16001|3.20000e+001, 1:16001|2.80000e+001, 1:16001|3.20000e+001
suite = 'PBO', funcId = 19, DIM = 100, algId = 'self_GA'
%
data_f19/IOHprofiler_f19_DIM100.dat, 1:100001|1.92000e+002, 1:100001|1.88000e+002, 
1:100001|1.80000e+002, 1:100001|1.76000e+002, 1:100001|1.76000e+002
\end{CodeInput}
\end{CodeChunk}
Note that the \iohana relies on the meta-data present in the info-files for its processing of associated data. Thus, it is crucial to ensure that these files are correct, especially when converting data from other formats into \iohpro or \emph{two-column} formats. 
The meta data is structured in the following ``three-line'' format (two examples of this ``three-line'' structure are provided in the example above), storing the high-level information on all instances of a tuple of (dimension, function).
\begin{itemize}
	\item \textbf{The first line} stores some meta-information of the experiment as (name, value) pairs. Note that such pairs are separated by commas and three names, \verb|funcId|, \verb|DIM| and \verb|algId| are \emph{case-sensitive} and \emph{mandatory}.
	\item \textbf{The second line} always starts with a single \verb|%|, indicating what follows this symbol should be the general comments from the user on this experiment. By default, it is left empty.
	\item \textbf{The third line} starts with the relative path to the actual data file, followed by the meta-information obtained on each instance, with the following format:
	$$\underbrace{1}_{\text{instance ID}}:\underbrace{1953125}_{\text{running time}}|\;\underbrace{5.59000e+02}_{\text{best-so-far f(x)}}$$
	By default, the data files (\verb|*.dat|, \verb|*.cdat|, \verb|*.tdat|, dots) are organized in the group of test functions, which are again stored in sub-folders with naming convention: \verb|data_[function ID]/|, e.g., \verb|data_f10/|. Moreover, when several dimensions are tested, the corresponding information above is written into the meta data one after the other. 
\end{itemize}

\paragraph{Raw-data} Despite the fact that different methods can be used to store data (resulting in four types of data file, {which also determines the extension, e.g.}  \verb|.dat| or \verb|.cdat|), the files take the same format, which is adapted from \texttt{csv} format to accommodate multiple runs/instances. An example of the structure of these files is shown below.
\begin{table*}[!htp]
\small
\centering
\ttfamily
\setlength{\tabcolsep}{4pt}
\begin{tabular}{rrrrrr}
"function evaluation" & "current f(x)" & "best-so-far f(x)" &  "parameter" & $\ldots$  \\
1 & +2.95000e+02 & +2.95000e+02 &  0.000000 & $\ldots$ \\
2 & +2.96000e+02 & +2.96000e+02 &  0.001600 & $\ldots$ \\
4 & +3.07000e+02 & +3.07000e+02 &  0.219200 & $\ldots$ \\
9 & +3.11000e+02 & +3.11000e+02 &  0.006400 & $\ldots$ \\
12 & +3.12000e+02 & +3.12000e+02 &  0.001600 & $\ldots$ \\
16 & +3.16000e+02 & +3.16000e+02 &  0.006400 & $\ldots$ \\
20 & +3.17000e+02 & +3.17000e+02 & 0.001600 & $\ldots$ \\
23 & +3.28000e+02 & +3.28000e+02 & 0.027200 & $\ldots$ \\
27 & +3.39000e+02 & +3.39000e+02 & 0.059200 & $\ldots$ \\
"function evaluation" & "current f(x)" & "best-so-far f(x)" & "parameter" & $\ldots$  \\
1  & +3.20000e+02 & +3.20000e+02 & 1.000000 & $\ldots$\\
24 & +3.44000e+02 & +3.44000e+02 & 2.000000 & $\ldots$\\
60 & +3.64000e+02 & +3.64000e+02 & 3.000000 & $\ldots$\\
"function evaluation" & "current f(x)" & "best-so-far f(x)" & "parameter" & $\ldots$  \\
	$\ldots$ & $\ldots$& $\ldots$ & $\ldots$ & $\ldots$ 
\end{tabular}
\end{table*}

Note that, each \emph{separation line} (line that starts with \verb|"function evaluation"|) serves as a separator among different independent runs of the same algorithm. Therefore, it is clear that the data block between two separation lines corresponds to a single run on a combination of dimension, function, and instance. 
In addition, a parameter value (named \verb|"parameter"|) is also tracked in this example and recording more parameter value is also facilitated (see below). 
Columns \verb|"current f(x)"| and \verb|"best-so-far f(x)"| stand for the current function value and the best one found so far, respectively. Here, \verb|"current f(x)"| stands for the function value observed when the corresponding number of function evaluation is performed while \verb|"best-so-far f(x)"| keeps track of the best function value observed since the beginning of one run. Only two columns, \verb|"function evaluation"| and \verb|"best-so-far f(x)"| are \textbf{mandatory} in this format. The \textbf{two-column} data format mentioned previously is to describe the minimal case where only those two columns are present in the raw data.

{In order to emulate the data format generated by IOHexperimenter, it is very important to pay attention to the following principles for the data format:}
\begin{itemize}
	\item The \emph{double quotation} (\verb|"|) in the separation line shall always be kept and it cannot be replace with single quotation (\verb|'|).
	\item The numbers in the record can either be written in the plain or scientific notation.
	\item To separate the columns, \emph{a single space or tab} can be used (only one of them should be used consistently in a single data file).
	\item If the performance data is tracked in the {improvement-based} scheme, where a row is written only if the \verb|"best-so-far f(x)"| is improved, the user \text{must} make sure that each block of records (as divided by the separation line) ends with the last function evaluation. {This allows the used budget to be extracted from the data-file when required.}
	\item Each data line should contain a complete record. Incomplete data lines will be dropped when loading the data into \iohana.
	\item The parameter columns, which record the state of (dynamic) internal parameters during the search, are fully customizable. The user can specify which parameter to track when running their algorithm using the IOHexperimenter. For more details on how to setup this parameter tracking in IOHexperimenter, please refer to our wiki page 	(\url{https://iohprofiler.github.io/IOHexp/Cpp/\#using-logger}).
	\item In case the quotation mark is needed in the parameter name, a single quotation (\verb|'|) should be used.
\end{itemize}
}

\end{document}